\pdfoutput=1 
\documentclass{article}

\usepackage[preprint,nonatbib]{neurips_2024}
\usepackage{microtype}
\usepackage{graphicx}
\usepackage{subfigure}
\usepackage{booktabs} 
\usepackage{bm,bbm}
\usepackage{multirow}
\usepackage{wrapfig}
\usepackage{amsmath, amsthm, amssymb, multirow, paralist}

\usepackage[utf8]{inputenc} 
\usepackage[T1]{fontenc}    
\usepackage{hyperref}       
\usepackage{url}            
\usepackage{booktabs}       
\usepackage{amsfonts}       
\usepackage{nicefrac}       
\usepackage{microtype}      
\usepackage{xcolor}         
\usepackage{amsmath}
\usepackage{amssymb}
\usepackage{mathtools}
\usepackage{amsthm}
\DeclareUnicodeCharacter{FF0C}{ }

\usepackage[capitalize,noabbrev]{cleveref}
\usepackage{stfloats}

\theoremstyle{plain}

\theoremstyle{definition}

\theoremstyle{remark}

\usepackage[textsize=tiny]{todonotes}

\title{Understanding the Role of Textual Prompts in LLM for Time Series Forecasting: an Adapter View}

\author{%
    Peisong Niu$^{*}$\quad Tian Zhou$^*$ \quad Xue Wang \quad Liang Sun \quad \textbf{Rong Jin} \\
    \texttt{\{tian.zt,niupeisong.nps,xue.w,liang.sun,jinrong.jr\}@alibaba-inc.com}%
}

\begin{document}

\maketitle

\let\thefootnote\relax\footnote{$*$ Equal contribution}
\begin{abstract}
    In the burgeoning domain of Large Language Models (LLMs), there is a growing interest in applying LLM to time series forecasting, with multiple studies focused on leveraging textual prompts to further enhance the predictive prowess. This study aims to understand how and why the integration of textual prompts into LLM can effectively improve the prediction accuracy of time series, which is not obvious at the glance,given the significant domain gap between texts and time series. Our extensive examination leads us to believe that (a) adding text prompts is roughly equivalent to introducing additional adapters, and (b) It is the introduction of learnable parameters—rather than textual information—that aligns the LLM with the time series forecasting task, ultimately enhancing prediction accuracy. Inspired by this discovery, we developed four adapters that explicitly address the gap between LLM and time series, and further improve the prediction accuracy. Overall, our work highlights how textual prompts enhance LLM accuracy in time series forecasting and suggests new avenues for continually improving LLM-based time series analysis. https://anonymous.4open.science/r/ExplicitAdapters4TS-1E6A.
\end{abstract}

\section{Introduction}\label{sec:introduction}
Time series analysis is a crucial area of research with profound implications for a breadth of real-world applications, such as energy, weather management, traffic forecasting, and economic trend analysis~\cite{wen2022robust}. Following their resounding triumphs in natural language processing (NLP)~\cite{vaswani2017attention} and computer vision (CV)~\cite{bao2022beit}, transformer-based models have recently become more popular in time series analysis, achieving promising results~\cite{zhou2022fedformer, wu2021autoformer, Patchformer, zhou2021informer, dcdetector, wen2022transformers}.

With the great success of foundational models for NLP~\cite{OpenAI2023GPT4TR, Touvron2023LLaMAOA} and CV~\cite{kirillov2023segany}, there is a growing interest of exploring these advanced models to time series tasks~\cite{garza2023timegpt1, rasul2024lagllama} due to their auto-regressive nature. Despite these efforts, time series forecasting still remains a challenging problem, mostly due to limited training data, significant data noise and outliers, and diverse dynamic patterns that vary significantly from domains to domains. 

An emerging group of approaches leveraging pre-trained language models for time series analysis has shown great potentials. They are backed by the findings that pre-trained transformers can offer universal representation and transferable capabilities~\cite{zhou2023onefitsall}. Even more interestingly, several studies showed that directing Large Language Models (LLMs) through textual prompts can further enhance forecasting performance for time series data, and as a result, multiple research efforts are devoted to designing effective textual prompts to improve the alignment between language model embeddings and time series data~\cite{jin2024timellm, xue2023promptcast, Jia2024GPT4MTSPL}.

\begin{figure}[h]
    \centering
    \includegraphics[width=1.0\columnwidth]{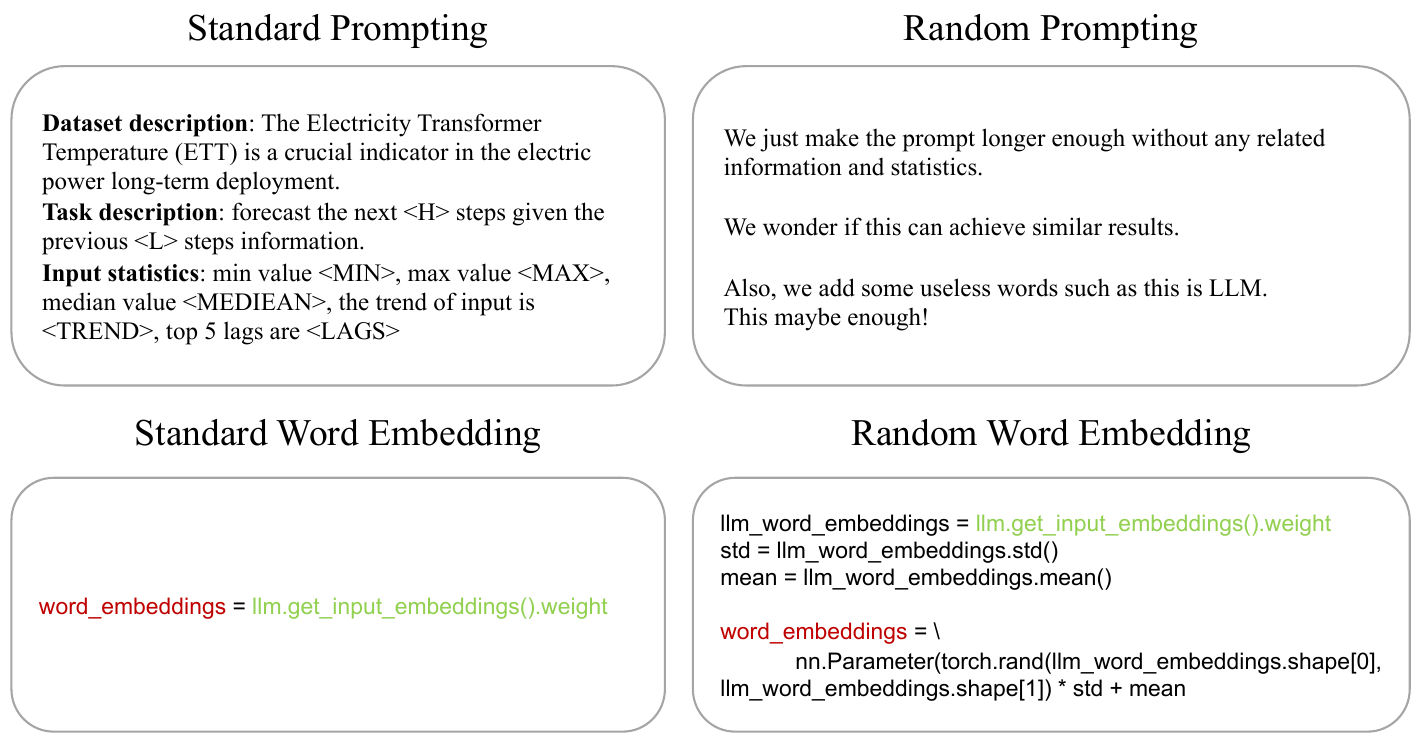}
    \caption{Illustration of random text test and random text embedding test.}
    \label{fig:llm_illustration}
\end{figure}

Although textual prompts are very effective in content generation and improving the reasoning ability of LLM, it is however unclear why they also effective for time series analysis. Due to the large domain gap between language and time series, it is difficult to envision how the semantic information in the textual prompts can help guide the LLM for better time series forecasting. To understand this question, we conduct an experiment that introduces random textual prompts and random word embeddings (see the examples in Figure ~\ref{fig:llm_illustration}) into LLM. Surprisingly, the experimental results (see details in Section 3.1) show that random prompts lead to similar performance improvements for time series forecasting as those well designed and semantically meaningful prompts. It thus leads to the main theme of this work: {\it What exactly is the role of textual prompts in LLM for time series forecasting ?}

Given that semantic information in textual prompts does not help explain the improvement, our alternative hypothesis is that these textual prompts may increase the model capacity and play a role similar to adapters that are commonly used in model fine tuning and domain adaption. We verify this hypothesis by conducting experiments with adapters introduced into LLM for time series forecasting and observing similar improvements as textual prompts (see details in Section 3.2 and 3.3). Subsequent experiments, however, revealed the limitations of using generic adapters in LLM for time series analysis: we quickly hit the performance plateau by continuously increasing the size of adapters. By carefully examining the results, we believe that the limitation is due to the lack of special designs of adapters to fill out the domain gap between language and time series, which can significantly restrain the power from LLM. Based on this discovery, we developed a set of specialized adapters, uniquely crafted to bridge the gap between language and time series. Our empirical studies indicate that these tailored adapters significantly outperform generic prompting and pet models in closing the modality gap, thus delivering substantial improvements in performance.


It is worth noting that despite our efforts in this work, \textbf{We are not suggesting that textual information cannot aid LLMs in time series analysis; however, its effectiveness has yet to be demonstrated.} We hope that our proposed simple testing method can be used for future text-prompted LLM-based models to validate their core designs. Additionally, our findings suggest that an adaptor is a viable alternative for achieving better accuracy in LLMs for time series analysis.


Outlining our core contributions, we present:

\begin{itemize}
    \item An extensive Investigation reveals the connection between adapters and textual prompts which used in LLM to help improve the performance of time series analysis.
    \item Two simple tests to determine whether the proposed model utilizes cross-modality information.
   
    \item Four specially designed adapters are proposed in this work to more effectively align LLMs with the unique inductive biases of time series data, significantly enhancing the performance of downstream tasks.
    
\end{itemize}

From here, the paper unfolds as follows: Section 2 surveys the burgeoning field of LLM applications in time series analysis. Section 3 
provides details of our investigation for the role of textual prompts in LLM to improve the performance of time series forecasting. Section 4 introduces our proposed explicit adapters tailored to align LLM with time series data. Section 5 provides comprehensive evaluations of these adapters within the broader context of time series analysis. And finally, Section 6 culminates with a discussion on our findings and contemplations of future research paths.
\section{Related Works}\label{sec:related_works}

In this section, we provide a short review of the literature in the areas of time series analysis with LLMs. These methods in the literature can be roughly categorized as implicit knowledge transfer and explicit linguistic guidance. The works aiming to build foundation models based on a substantial amount of time series data are not mentioned here.

\textbf{Implicit Knowledge Transfer from Pre-trained LLMs}
Transformer excels in its capability to simultaneously process data from various modalities and domains through self-attention. This has sparked exploration into the transferability of pre-trained models across different modalities. For time series, OFA~\cite{zhou2023onefitsall} pioneered the exploration of transferring knowledge from pre-trained GPT-2 models, showcasing the transformers' ability to offer a universal representation. 
In order to achieve a better alignment between the embeddings of time series and linguistic data, LLM4TS~\cite{Chang2023LLM4TSAP} proposes a preliminary stage, different from the linear probing approach adopted by OFA, where LLMs are aligned with the nuances of time series data.
TEST~\cite{Sun2023TESTTP} introduces a data-centric paradigm, TS-for-LLM, which involves converting time series data into a format that is compatible with language, enabling effective utilization of LLMs in time series analysis.
Both of these approaches involve the implicit transfer of pre-trained knowledge from linguistic data, either through fine-tuning on time series data or by adjusting the embedding distribution.

\textbf{Explicit Linguistic Guidance with LLMs}
Apart from exploration the transfer ability of LLMs, it is truly remarkable how linguistic information can explicitly guide models for time series analysis. 
PromptCast~\cite{xue2023promptcast} transforms the numerical input and output into prompts and making it possible to directly apply LLMs for forecasting. However, the zero-shot setting heavily relies on the mathematical reasoning ability of LLMs, which remains a significant challenge, and there is a lack of comprehensive reviews regarding the selection strategies of LLMs.
GPT4MTS~\cite{Jia_Wang_Zheng_Cao_Liu_2024} leverages the combination of numerical data with linear embedding and textual information with BERT embedding~\cite{Bert/NAACL/Jacob} simultaneously, surpassing OFA without explicit prompts. 
Time-LLM~\cite{jin2024timellm} introduces Prompt-as-Prefix and reprograms the input time series with text prototypes. With LLaMA~\cite{rasul2024lagllama}, Time-LLM significantly outperforms OFA and achieves SOTA.
Yet, both GPT4MTS and Time-LLM lack a comprehensive analysis of prompt selection and the impact of linguistic semantics on performance.  Additionally, these models could benefit from the introduction of more learnable parameters.

\section{Understanding Textual Prompts in LLM for Time Series Analysis }\label{sec:explore_linguistic}

In this section, we present a series of investigations proposing that the success of recent methods using textual prompts with Large Language Models (LLMs) for time series forecasting is primarily due to the function of implicit adapters.

\subsection{Is Semantic Meaning of Textual Prompts Beneficial for Time Series Forecasting ?}


To assess whether semantic information in textual prompts can enhance time series forecasting by large language models (LLMs), we took inspiration from recent studies that incorporate both texts and word embeddings in prompts. We devised two sets of tests: the Random Textual Test, which uses random text as prompts, and the Random Word Embedding Test, which employs random text embeddings. Examples of both tests are shown in Figure.~\ref{fig:llm_illustration}. 
\begin{enumerate}
    \item \textbf{Random text Test}: The input prompt consists of meaningless sentences that are unrelated to the time series.
    \item \textbf{Random text Embedding Test}: The time series embeddings are aligned with random learnable values.
\end{enumerate}
Compared to the well designed prompts, these random prompts do not have special meaning and thus will help verify our hypothesis. 


We compare multiple LLMs, including GPT-2~\cite{gpt2-2019} with 12 layers and LLaMA~\cite{Touvron2023LLaMAOA} with 32 or 8 layers. Due to memory constraints, for LLaMA(32), we use 4-bit quantization.
The performance on ETTh1 96 is shown in Table~\ref{tab:llm_illustration}. 
In the case of GPT-2(12) and LLaMA(8), utilizing random prompting and random word embedding can achieve comparable performance to that of standard textual prompts. Notably, LLaMA(8) with random prompting even exhibits a slight improvement. 
This result clearly shows that the semantic meaning of textual prompts is not the key driving force for improving the prediction accuracy of time series, leading us to examine the second hypothesis, i.e., the introduction of textual prompts essentially increases the model size and may play a role similar to adapters that are commonly used for model fine tuning and domain adaption. 


\begin{table*}[h]
\caption{Performance (MSE) with various prompting and word embedding. S, RP and RW represent Standard, Random Prompting and Random Word Embedding respectively. Due to the memory limitation, * represents the input length of 96. Also, it is hard to perform experiments with LLaMA(32) on most datasets. Therefore, the results for ETTh1 and ETTh2 are provided in the Appendix \ref{app:llama32}. '-' represents out of memory.
}
\vskip -0.10in
\label{tab:llm_illustration}
\begin{center}
\scalebox{0.85}{
\begin{tabular}{cc|cc|cc|cc|cc|cc}

\toprule


& & \multicolumn{2}{c|}{ETTh1} & \multicolumn{2}{c|}{ETTh2} & \multicolumn{2}{c|}{Weather} & \multicolumn{2}{c|}{Traffic} & \multicolumn{2}{c}{ECL} \\

& & 96 & 192 & 96 & 192 & 96 & 192 & 96 & 192 & 96 & 192 \\

\midrule

& S & 0.385 & 0.419 & 0.306 & 0.332 & 0.152 & 0.200 & 0.355 & 0.398 & 0.140 & 
 0.158 \\
GPT-2(12)& RP & 0.385 & 0.409 & 0.303 & 0.333 & 0.146 & 0.194 & 0.354 & 0.402 & 0.134 & 0.153 \\
& RW & 0.387 & 0.413 & 0.309 & 0.335 & 0.152 & 0.199 & 0.354 & 0.398 & 0.141 & 0.154 \\

\midrule

& S & 0.389 & 0.412 & 0.297 & 0.329 & *0.187 & *0.239 & -&- & - & - \\
LLaMA(8)& RP & 0.384 & 0.411 & 0.304 & 0.333 & *0.190 & *0.240 & -&- &- &- \\
& RW & 0.393 & 0.411 & 0.296 & 0.335 & *0.197 & *0.249 &- &- & -&- \\





\bottomrule

\end{tabular}
}
\end{center}
\vskip -0.20in
\end{table*}

\subsection{From Textual Prompts to Adapter Methods}


To understand if textual prompts in LLM play a similar role as adapters for time series forecasting, we conduct experiments with various adapters or methods for efficient parameter fine tuning, including LoRA~\cite{hu2022lora} and Prefix Tuning~\cite{li2021prefixtuning}. All the experiments are conducted using the OFA framework~\cite{zhou2023onefitsall}, a generic framework for directly adopting LLM for time series forecasting. In Table~\ref{tab:adapters_test}, compared to the baseline OFA without any adapter, we observe an improvement in prediction error by using different adapters. By comparing these results to the that for textual prompts in Table~\ref{tab:llm_illustration}, we observe that explicit adapters yield similar improvements as textual prompts for LLM-based time series analysis, indicating that textual prompts may play a similar role as adapters.  


\begin{table}[h]
\caption{Various methods on ETTh1 and ETTm1 (MSE).}
\vskip -0.10in
\label{tab:adapters_test}
\begin{center}
\scalebox{0.8}{
\setlength\tabcolsep{3pt}
\begin{tabular}{c|ccc|c}

\toprule
Datasets & TimeLLM-GPT2(6) & Prefix Tuning & LoRA & OFA \\

\midrule
  
ETTh1 96 &0.394&0.368&0.374&0.376 \\
ETTh1 192 &0.427&0.404&0.410&0.416  \\
\midrule
ETTm1 96 &0.311&0.292&0.294&0.292  \\
ETTm1 192 &0.342&0.330&0.329&0.332  \\

\bottomrule

\end{tabular}
}
\end{center}
\vskip -0.15in
\end{table}



\subsection{Incorporating Time Series-Specific Information into Adapters}

One shortcoming of the generic adapters is that by increasing the size of adapters, they fail to continuously improve the prediction accuracy, as shown in Table~\ref{tab:exp_param}. 
Given the role of adapters is to bridge the gap between language and time series, we find it is important to design special adapters that explicitly address the challenges faced by time series data, i.e., temporal patterns that are dynamically changing, correlation across different channels, and outlier data points. To verify this hypothesis, we explore time series-specific adapters to see if they can provide additional gains compared to the generic ones, where more discussions can be found in Section \ref{sec:methodology}. 
Table \ref{tab:adapters_design} showcases OFA equipped with time-series specific adapters, where temporal and channel adapters explicitly introduce time series-specific information and both outperform vanilla adapter with equivalent learnable parameters, guiding the design of adapters.

\begin{table}[ht]
\caption{Learnable parameter comparison (MSE).}
\label{tab:exp_param}
\vskip -0.10in
\begin{center}

\scalebox{0.75}{
\setlength\tabcolsep{3pt}
\begin{tabular}{c|c|ccc|cccc}
\toprule

\multirow{2}{*}{Methods} & \multirow{2}{*}{OFA}

& \multicolumn{3}{c|}{Prefix Tuning(\#Prefix)} & \multicolumn{4}{c}{LoRA(Rank)} \\

& & 1 & 5 & 20 & 8 & 16 & 64 & 128 \\

\midrule
ETTh1 96 &0.376 & 0.368 & 0.369 & 0.368 & 0.376 & 0.374 & 0.372 & 0.373 \\
ETTh2 96 &0.285 & 0.275 & 0.277 &0.276 & 0.278 & 0.281 & 0.284 & 0.283 \\
\midrule
\#Additinal Params& 0 & 768 & 3,840 & 15,360 & 258,048 & 516,096 & 2,064,384 & 4,128,768\\

\bottomrule
\end{tabular}%
}
\end{center}
\vskip -0.10in
\end{table}


\begin{table}[ht]
\caption{The performance (MSE) of OFA with various adapters on ETTh1 and ETTh2.}
\vskip -0.10in
\label{tab:adapters_design}
\begin{center}
\scalebox{0.75}{
\setlength\tabcolsep{3pt}
\begin{tabular}{c|c|ccc}

\toprule
Datasets & OFA & +Vanilla (Temporal) Adapter & +Channel Adapter 
 & +Frequency Adapter \\

\midrule

ETTh1 96 & 0.376 &0.375 & 0.371 & 0.367\\
ETTh1 192 & 0.416 &0.413&0.409&0.406\\
\midrule
ETTh2 96 & 0.285 &0.280 &0.279& 0.269\\
ETTh2 192 & 0.354 & 0.348 & 0.344 & 0.334\\

\bottomrule

\end{tabular}
}
\end{center}
\vskip -0.20in
\end{table}


\section{Developing Explicit Adapters for Large Language Models in Time Series Analysis}\label{sec:methodology}

\subsection{Overview Methodology}
\begin{figure*}[ht]
    \centering
    \includegraphics[width=0.8\columnwidth]{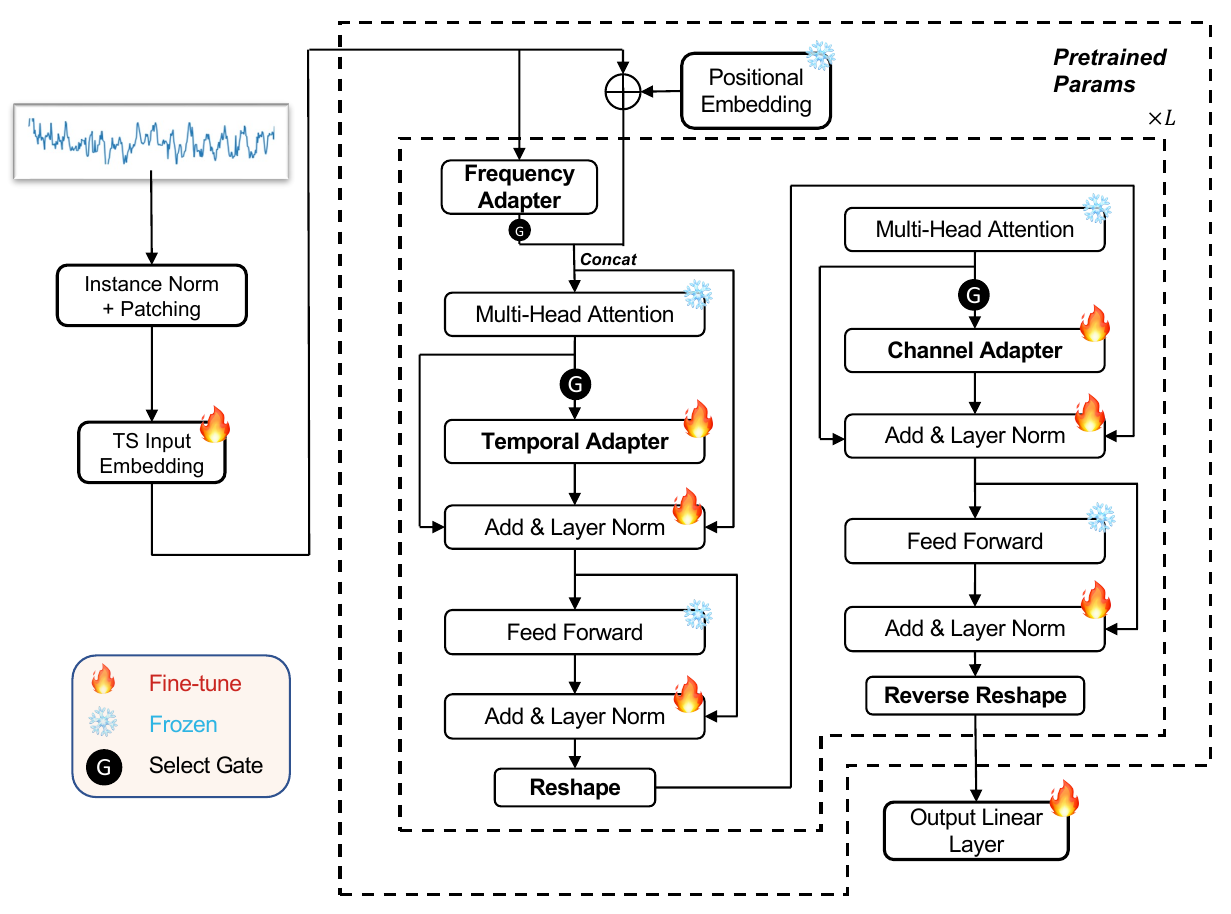}
    \caption{Model architecture: The self-attention and feedforward layers in the transformer blocks, along with the positional embeddings, are fixed. Designed adapters are integrated into each transformer block of the pre-trained model and fine-tuned during the training process.}
    \label{fig:structure_2}
    \vskip -0.10in
\end{figure*}

Section \ref{sec:explore_linguistic} has validated the effectiveness of explicit adapters. To incorporate time series-specific information, we design various adapters into OFA.
The overall architecture of our proposed method is shown in Figure \ref{fig:structure_2}. After undergoing instance normalization and patching, the primary tokens for transformers are mapped by the TS input embedding module. 
Finally, the representation of last block is applied into various downstream tasks.

\subsection{Instance Normalization, Patching and Frozen Pre-trained Block}

\textbf{Instance Normalization} We integrate a straightforward data normalization block, non-affine reverse instance normalization~\cite{kim2022reversible}, by normalizing the input time series using the mean and variance, and subsequently adding them back to the output.



\textbf{Patching} Similar to PatchTST~
\cite{Patchformer}, we aggregate adjacent time steps to form a single patch-based token for transformer.


\textbf{Frozen Pre-trained Block} Same as OFA \cite{zhou2023onefitsall}, we retain the positional embedding layers and self-attention blocks from the pre-trained models. Also, the self-attention blocks and feedforward layers are frozen while fine-tuning.



\subsection{Adapters}

To more effectively integrate specific knowledge from time series into the language model, we propose various adaptors, including temporal, channel, frequency and anomaly adaptors for incorporating the relevant information. Model design analysis can be found in Appendix \ref{app:model_selection}. We must emphasize that our proposed adapters are both straightforward and highly effective. \textbf{We've deliberately kept their structure simple, incorporating only minimal bias to enhance the model's performance with negligible additional computational overhead.}

\textbf{Temporal \& Channel Adapters}
Both OFA~\cite{zhou2023onefitsall} and PatchTST~\cite{Patchformer} employ a channel-independent approach, concentrating primarily on modeling temporal correlations. While this method can significantly reduce the risk of overfitting, it also neglects cross-channel information. In real-world tasks, we often have access to a large number of channels or covariates in time series data, which contain valuable information that can enhance forecasting performance. However, directly modeling the pairwise correlations among channels tends to overfit the training data, resulting in poor generalization performance. 

\begin{figure}[htbp]
    \centering
    \begin{minipage}[t]{0.42\textwidth}
        \centering
        \includegraphics[width=1\columnwidth]{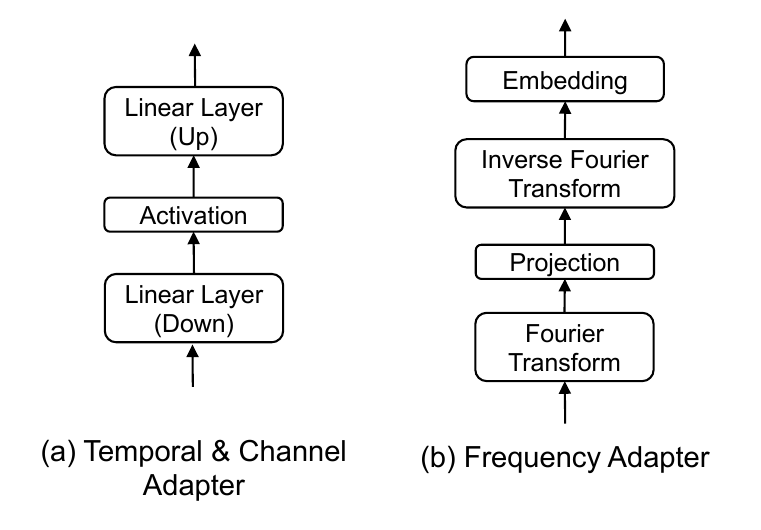}
        \caption{Structures of various adapters.}
        \label{fig:structure_adapter}
    \end{minipage}
    \begin{minipage}[t]{0.4\textwidth}
        \centering
        \includegraphics[width=1\columnwidth]{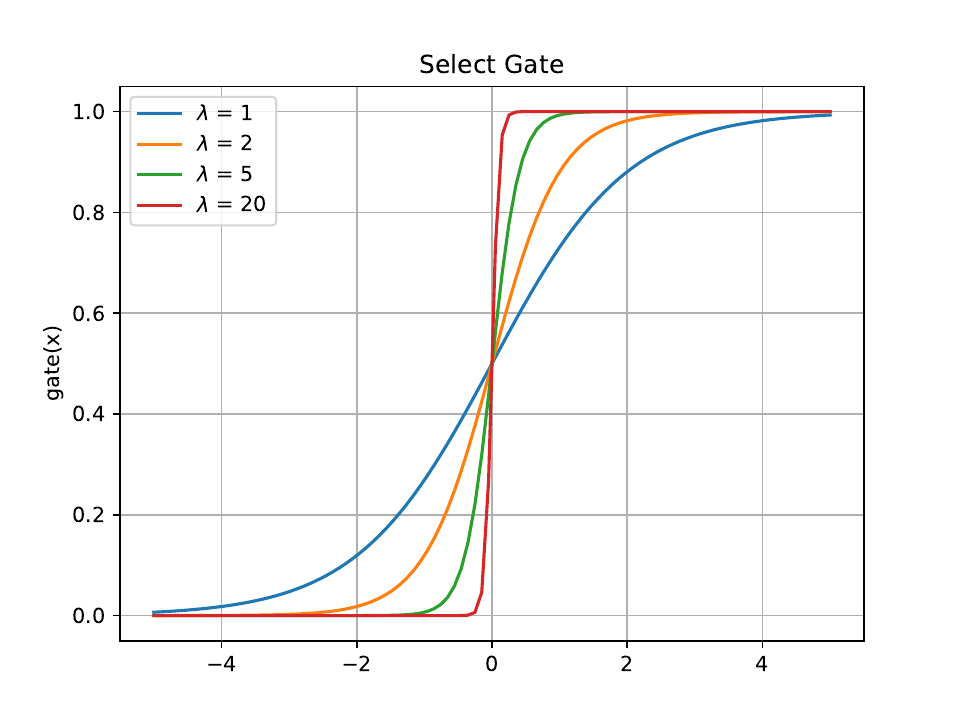}
        \caption{Select gate with various scaling factors.}
        \label{fig:select_gate}
    \end{minipage}
    \vskip -0.25in
\end{figure}

To address this challenge, we introduce a ``compressor'' like channel adaptor with a bottleneck structure, as shown in Figure~\ref{fig:structure_adapter} (a). Using the bottleneck structure, we are able to capture the channel correlation through the hidden space without suffering from the overfitting. A similar structure is also used for temporal adaptor. For each transformer block, we duplicate the self-attention module and insert different adapters after the multi-head attention modules by reshaping the feature dimension.




\textbf{Frequency Adapters}
The time series data can also be modeled in the frequency domain where seasonal and global information can be captured easily. Motivated by the success of frequency modeling~\cite{zhou2022fedformer,zhang2022TFAD}, we design a frequency adapter, as illustrated in Figure~\ref{fig:structure_adapter}(b). Specifically, for patched input time series, we initially convert it from the time domain within each patch to the frequency domain through Fast Fourier Transform (FFT). Subsequently, we perform a projection before applying inverse FFT. This adapter is designed to incorporate frequency domain bias modeling using the simplest structure.




\textbf{Select Gate}
The information captured in time series varies depending on the domain. Thus to better adapt to various datasets, an adaptable selection mechanism to better concentrate on specific adapter plays a pivotal role in enhancing overall performance. As in Eq. \ref{eq3} and Figure \ref{fig:select_gate},
Given that various signals or hidden variables may exhibit distinct biases, leading to different preferences for our designed adapters, we implement a scaled sigmoid as an adaptive gating mechanism. This is placed before each adapter to modulate the strength of each one within each block. 
\begin{equation}
    \label{eq3}
    {\rm gate}(g) = \sigma(\lambda g) = \frac{1}{1 + e^{-\lambda g}},
\end{equation}
where $g$ is a learnable parameter and $\lambda$ represents the scaling factor.


\begin{figure}[h]
    \centering
    \includegraphics[width=0.6\columnwidth]{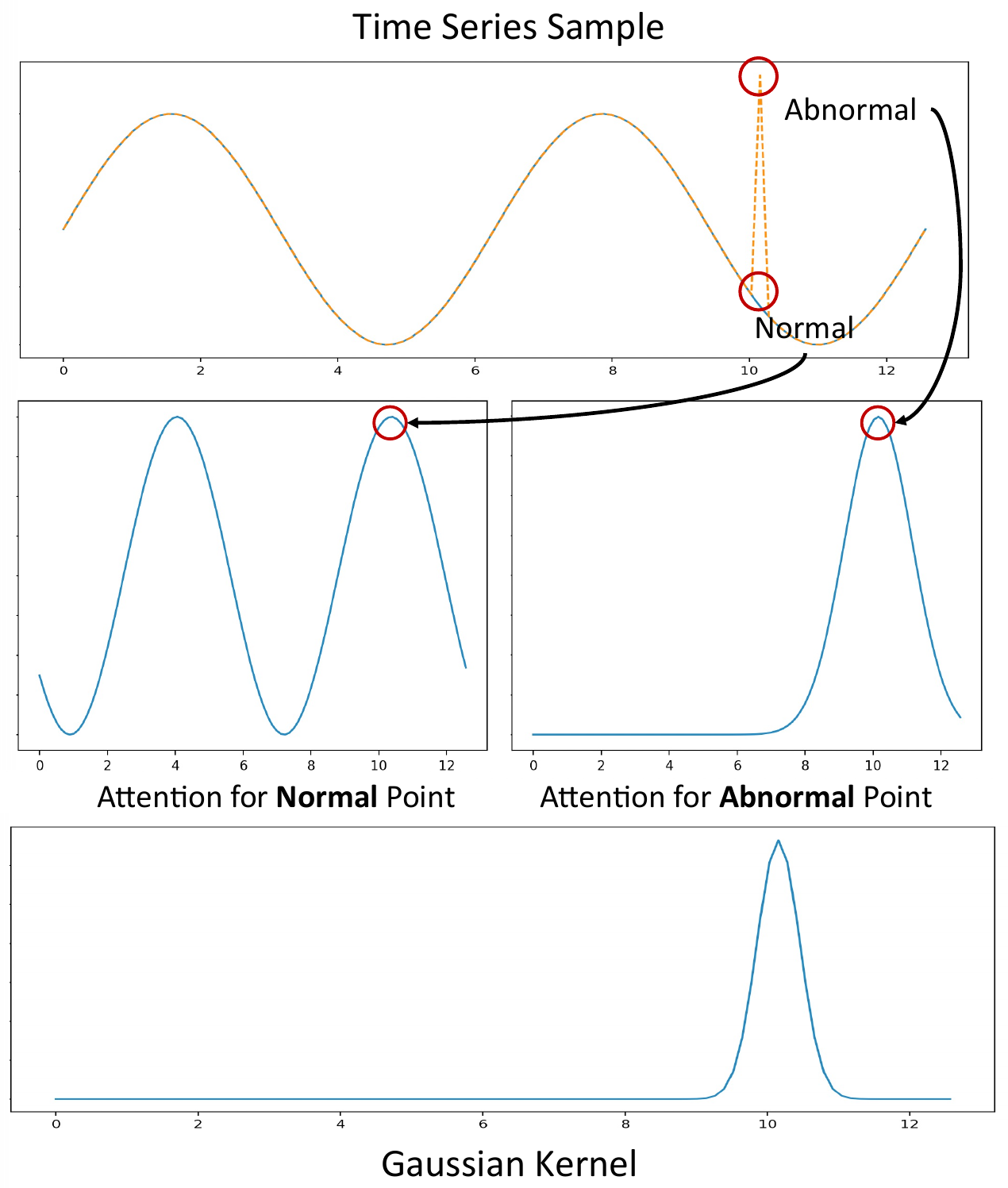}
    \caption{Attention and gaussian kernel anomaly adaptor for normal and abnormal points.}
    \label{fig:anomaly_adapter}
    \vskip -0.1in
\end{figure}

\textbf{Anomaly Adapter}
OFA~\cite{zhou2023onefitsall} and TimesNet~\cite{timesnet} perform anomaly detection based on the reconstruction loss between ground truth and predicted values. However, they still fall short of matching the performance of SOTA methods specifically designed for time series anomaly detection, largely due to their reliance on a purely regression-based detection framework.~\cite{xu2021anomaly, dcdetector}. 
Figure \ref{fig:anomaly_adapter} shows the design of anomaly adapter.In order to capture this property for more effective time series anomaly detection, we develop an anomaly adapter by capturing the contrastive bias through the self-attention matrix.

In particular, normal patterns tend to repeat themselves over time, 
while such periodical patterns will be absent from the self-attention matrix for anomaly signals. By treating numbers in each row of the self-attention matrix as a distribution after appropriately normalization, we can measure the difference by the KL divergence. 

Specifically, for the $l^{th}$ layer, we calculate the distribution difference between attention $A_l$ and the outputs of anomaly adapter $A^{anomaly}_l$:
\begin{equation}
    \label{eq4}
    loss^{discrepancy}_l = {\rm KL}(\hat{A}_l||A^{anomaly}_l)
\end{equation}
\begin{equation}
    \label{eq5}
    \hat{A}_l = \frac{1}{2}[A + A^T - {\rm diag}(A)]
\end{equation}
\begin{equation}
    \label{eq6}
    A^{anomaly}_{l_{i, j}} = [\frac{1}{\sqrt{2\pi}\sigma_i}{\rm exp}(-\frac{{\rm dis}(i, j)}{2\sigma_i^2})],
\end{equation}
where $\sigma_i$ is learnable and ${\rm dis}(i, j)$ represents the distance between $i^{th}$ and $j^{th}$ tokens. As shown in Table \ref{tab:anomaly_sota}, it significantly improves performance.

\section{Experiments}\label{sec:ts:task}

\subsection{Explicit Adapters Versus Implicit Adapters}
The Time-LLM framework stands out for its significant contributions to time series forecasting using LLMs, especially with its innovative application of both textual prompts and text embedding alignment. Being a seminal work in this area, it serves as a primary example of implicit adapter utilization in our study. 

Table \ref{tab:exp_imp} gives a thourough comparison between GPT2-adapter and Time-LLM. 
GPT2-adapter(6) consistently outperforms Time-LLM with LLMs of similar parameter size.
Furthermore, it is noteworthy that GPT2-adapter(6) matches the performance of Time-LLM-LLaMA(32) despite relying on a backbone that is \textbf{100 times smaller}.
This suggests that explicit adapters can also yield comparable improvements. 
Furthermore, it underscores the necessity for further exploration into how to effectively utilize semantic textual information in time series analysis.

\begin{table*}[ht]
\caption{Long-term forecasting task. All the results are averaged from 4 different prediction lengths, that is $O \in \{96, 192, 336, 720\}$. A lower MSE indicates better performance. \textbf{Black}: best, \underline{Underline}: second best.}
\label{tab:long_term}
\begin{center}

\scalebox{0.95}{
\setlength\tabcolsep{3pt}
\begin{tabular}{c|cc|cc|cc|cc|cc|cc|cc}
\toprule

\multirow{2}{*}{Methods}
&\multicolumn{2}{c|}{GPT2(6)-adapter}&\multicolumn{2}{c|}{OFA}&\multicolumn{2}{c|}{DLinear}&\multicolumn{2}{c|}{PatchTST}&\multicolumn{2}{c|}{TimesNet}&\multicolumn{2}{c|}{FEDformer}&\multicolumn{2}{c}{Autoformer} \\
& MSE  & MAE & MSE & MAE& MSE & MAE& MSE   & MAE& MSE  & MAE& MSE  & MAE& MSE  & MAE \\
\midrule

Weather &{\bf0.219}&{\bf0.250}&0.237&0.270&0.248&0.300&\underline{0.225}&\underline{0.264}&0.259&0.287&0.309&0.360&0.338&0.382\\

ETTh1&{\bf0.406}&\underline{0.427}&0.427&\textbf{0.426}&0.422&0.437&\underline{0.413}&0.430&0.458&0.450&0.440&0.460&0.496&0.487\\

ETTh2&\underline{0.338}&\underline{0.385}&0.354&0.394&0.431&0.446&\textbf{0.330}&\textbf{0.379}&0.414&0.427&0.437&0.449&0.450&0.459\\

ETTm1&{\bf0.348}&{\bf0.371}&0.352&0.383&0.357&\underline{0.378}&\underline{0.351}&0.380&0.400&0.406&0.448&0.452&0.588&0.517 \\

ETTm2&{\bf0.247}&{\bf0.307}&0.266&0.326&0.267&0.333&\underline{0.255}&\underline{0.315}&0.291&0.333&0.305&0.349&0.327&0.371 \\


ECL&\textbf{0.159}&\textbf{0.252}&0.167&0.263&0.166&0.263&\underline{0.161}&\textbf{0.252}&0.192&0.295&0.214&0.327&0.227&0.338\\

Traffic&\underline{0.397}&{\bf0.257}&0.414&0.294&0.433&0.295&\textbf{0.390}&\underline{0.263}&0.620&0.336&0.610&0.376&0.628&0.379\\
\midrule
Avg. &{\bf0.302}&{\bf0.321}&0.316&0.336&0.332&0.350&\underline{0.304}&\underline{0.326}&0.376&0.362&0.394&0.396&0.436&0.419\\

\bottomrule
\end{tabular}%
}

\end{center}
\end{table*}
\begin{table*}[h]
\caption{Short-term forecasting task on M4. The prediction lengths are in [6, 48] and results are weighted averaged from several datasets under different sample intervals. A lower SMAPE value indicates better performance. \textbf{Black}: best, \underline{Underline}: second best.}
\label{tab:short_term}
\begin{center}
\resizebox{0.9\textwidth}{!}{%
\begin{tabular}{c|cccccccccc}
\toprule

Methods&GPT2(6)-adapter&OFA&TimesNet&PatchTST&N-HiTS&N-BEATS& DLinear &FEDformer &Autoformer \\
\midrule

SMAPE &{\bf11.713}&11.991 &\underline{11.829}&12.059& 11.927& 11.851& 13.639 &12.840 &12.909  \\
MASE &{\bf1.572}& 1.600 & \underline{1.585}&1.623 & 1.613 & 1.599  &2.095 &1.701 &1.771  \\
OWA &{\bf0.858}&0.879 & \underline{0.867}&0.890 &0.881 &0.870  &1.051 &0.952 &0.972 \\

\bottomrule

\end{tabular}%
}
\end{center}
\end{table*}
\begin{table*}[ht]
\caption{Anomaly detection task. We calculate the F1-score (as \%) for each dataset. A higher F1 indicates better performance. \textbf{Black}: best, \underline{Underline}: second best.}
\vskip -0.10in
\label{tab:anomaly_sota}
\begin{center}
\resizebox{0.9\textwidth}{!}{%
\begin{tabular}{c|ccccccccc}
\toprule

Methods&GPT2(6)-adapter&OFA&DCdetector&Anomaly&TimeNet&THOC&InterFusion&OmniAnomaly&BeatGAN \\
\midrule
SMD&\underline{89.99}&86.89&87.18&\textbf{92.33}&84.61&84.99&86.22&85.22&78.10 \\

MSL&\underline{93.60}&82.45&\textbf{96.60}&93.59&81.84&89.69&86.62&87.67&87.53 \\

SMAP&\underline{96.88}&72.88&\textbf{97.02}&96.69&69.39&90.68&89.14&86.92&69.60 \\

SWaT&\textbf{98.08}&94.23&\underline{96.33}&94.07&93.02&85.13&83.01&82.83&79.92 \\

PSM&\textbf{98.22}&97.13&\underline{97.94}&97.89&97.34&89.54&83.52&80.83&92.83 \\
\midrule

Avg. &\textbf{95.35}&86.72&\underline{95.01} &94.91&85.24&88.01&85.70&84.69&81.60 \\
\midrule
\end{tabular}%
}
%
\end{center}
\vskip -0.10in
\end{table*}

\begin{table}[h]
\caption{Comparison between explicit adapters (GPT2-adapter) and implicit adapters (TimeLLM) with similar parameter size on ETTh1 and ETTm1. {\bf Black}: best, \underline{Underline}: second best.}
\label{tab:exp_imp}
\begin{center}
\vskip -0.10in
\scalebox{0.95}{
\setlength\tabcolsep{3pt}
\begin{tabular}{c|cccc}
\toprule

Methods & ETTh1 96 & ETTh1 192 & ETTm1 96 & ETTm1 192 \\
\midrule
TimeLLM-GPT2(6) & 0.394 & 0.427 & 0.311 & 0.342 \\
TimeLLM-GPT2(12) & 0.385 & 0.419 & 0.306 & 0.332 \\
TimeLLM-LLaMA(8) & 0.389 & 0.412 & 0.297 & \underline{0.329} \\
TimeLLM-LLaMA(32) & \bf{0.362} & \bf{0.398} & \bf{0.272} & \bf{0.310} \\
\midrule
GPT2-adapter(6) & \underline{0.367} & \underline{0.406} & \underline{0.292} & 0.330 \\

\bottomrule
\end{tabular}%
}
\end{center}
\vskip -0.15in
\end{table}

\subsection{Adapt Explicit Adapters to General Time Series Analysis}

To ensure a fair comparison, we adhere to the experimental settings of OFA \cite{zhou2023onefitsall}. The results demonstrate that the proposed methods excels in various downstream tasks with adapters, including time series long/short-term forecasting, anomaly detection, classification,imputation, few-shot and zero-shot forecasting.

\begin{table*}[ht]
\caption{Results for the classification task. $\ast$. in the Transformers indicates the name of $\ast$former. A higher accuracy indicates better performance. \textbf{Black}: best, \underline{Underline}: second best.}
\label{tab:classification}
\centering
\scalebox{0.8}{
\begin{tabular}{c|c|c|c|ccccc|c|c|cc}
\toprule
\multirow{2}{*}{Methods} &
Classical & RNN & \multirow{2}{*}{TCN} & \multicolumn{5}{c|}{Transformers} & \multirow{2}{*}{DLinear} & \multirow{2}{*}{TimesNet} & \multirow{2}{*}{OFA} & \multirow{2}{*}{GPT2(6)-adapter} \\
& Rocket & LSTNet & & Auto. & Station. & FED. & ETS. & Flow. & & & & \\
\midrule
EthanolConcentration & 45.2 & 39.9 & 28.9 & 31.6 & 32.7 & 31.2 & 28.1 & 33.8 & 32.6 & \textbf{35.7} & 34.2 & \underline{35.0} \\
FaceDetection & 64.7 & 65.7 & 52.8 & 68.4 & 68.0 & 66.0 & 66.3 & 67.6 & 68.0 & 68.6 & \underline{69.2} & \textbf{69.7} \\
Handwriting & 58.8 & 25.8 & 53.3 & 36.7 & 31.6 & 28.0 & 32.5 & \textbf{33.8} & 27.0 & 32.1 & \underline{32.7} & 32.2 \\
Heartbeat & 75.6 & 77.1 & 75.6 & 74.6 & 73.7 & 73.7 & 71.2 & 77.6 & 75.1 & \underline{78.0} & 77.2 & \textbf{79.5} \\
JapaneseVowels & 96.2 & 98.1 & 98.9 & 96.2 & 99.2 & 98.4 & 95.9 & \textbf{98.9} & 96.2 & 98.4 & \underline{98.6} & 98.1 \\
PEMS-SF & 75.1 & 86.7 & 86.1 & 82.7 & 87.3 & 80.9 & 86.0 & 83.8 & 75.1 & \textbf{89.6} & \underline{87.9} & 86.1 \\
SelfRegulationSCP1 & 90.8 & 84.0 & 84.6 & 84.0 & 89.4 & 88.7 & 89.6 & 92.5 & 87.3 & 91.8 & \textbf{93.2} & \textbf{93.2} \\
SelfRegulationSCP2 & 53.3 & 52.8 & 55.6 & 50.6 & 57.2 & 54.4 & 55.0 & 56.1 & 50.5 & 57.2 & \underline{59.4} & \textbf{60.6} \\
SpokenArabicDigits & 71.2 & \textbf{100.0} & 95.6 & \textbf{100.0} & \textbf{100.0} & \textbf{100.0} & \textbf{100.0} & 98.8 & 81.4 & 99.0 & 99.2 & 98.7 \\
UWaveGestureLibrary & \textbf{94.4} & 87.8 & \underline{88.4} & 85.9 & 87.5 & 85.3 & 85.0 & 86.6 & 82.1 & 85.3 & 88.1 & 87.5 \\
\midrule
Average & 72.5 & 71.8 & 70.3 & 71.1 & 72.7 & 70.7 & 71.0 & 73.0 & 67.5 & 73.6 & \underline{74.0} & \textbf{74.1} \\
\bottomrule
\end{tabular}
}
\vskip 0in
\end{table*}

\begin{table*}[ht]
\caption{ Imputation task. We randomly mask {12.5\%, 25\%, 37.5\%, 50\%} time points of 96-length time series. 
A lower MSE indicates better performance. \textbf{Black}: best, \underline{Underline}: second best.}
\vskip -0.10in
\label{tab:imputation}
\begin{center}
\resizebox{0.9\textwidth}{!}{%
\setlength\tabcolsep{3pt}
\begin{tabular}{c|cc|cc|cc|cc|cc|cc|cc|cc}
\toprule

\multirow{2}{*}{Methods} &\multicolumn{2}{c|}{GPT2(3)-adapter}
&\multicolumn{2}{c|}{OFA} & \multicolumn{2}{c|}{TimesNet}& \multicolumn{2}{c|}{PatchTST}&\multicolumn{2}{c|}{DLinear}&\multicolumn{2}{c|}{FEDformer}&\multicolumn{2}{c|}{Stationary}&\multicolumn{2}{c}{Autoformer} \\
&MSE&MAE&MSE&MAE&MSE&MAE&MSE&MAE&MSE&MAE&MSE&MAE&MSE&MAE&MSE&MAE \\

\midrule
ETTm1&{\bf0.025}&{\bf0.105}&\underline{0.028}&{\bf0.105}&0.027&0.107&0.047&0.140&0.093&0.206&0.062&0.177&0.036&0.126&0.051&0.150 \\

ETTm2&\underline{0.022}&0.091&{\bf0.021}&{\bf0.084}&\underline{0.022}&\underline{0.088}&0.029&0.102&0.096&0.208&0.101&0.215&0.026&0.099&0.029&0.105 \\

ETTh1&{\bf0.068}&\underline{0.174}&\underline{0.069}&{\bf0.173}&0.078&0.187&0.115&0.224&0.201&0.306&0.117&0.246&0.094&0.201&0.103&0.214 \\

ETTh2&{\bf0.045}&{\bf0.136}&\underline{0.048}&\underline{0.141}&0.049&0.146&0.065&0.163&0.142&0.259&0.163&0.279&0.053&0.152&0.055&0.156 \\

ECL &{\bf0.080}&{\bf0.196}&\underline{0.090}&\underline{0.207}&0.092&0.210&0.072&0.183&0.132&0.260&0.130&0.259&0.100&0.218&0.101&0.225 \\

Weather&{\bf0.030}&\underline{0.055}&0.031&0.056&{\bf0.030}&{\bf0.054}&0.060&0.144&0.052&0.110&0.099&0.203&0.032&0.059&0.031&0.057 \\
\midrule

Avg.&{\bf0.045}&{\bf0.126}&\underline{0.048}&\underline{0.128}&0.050&0.132&0.064&0.159&0.119&0.224&0.112&0.229&0.056&0.142&0.061&0.151 \\
\bottomrule

\end{tabular}%
}
\end{center}
\vskip -0.10in
\end{table*}
\begin{table*}[!ht]
\caption{Few-shot learning results on 10\% data. We use prediction length $O \in \{96, 192, 336, 720\}$. A lower MSE indicates better performance. 
\textbf{Black}: best, \underline{Underline}: second best.}
\vskip -0.10in
\setlength\tabcolsep{3pt}
\label{tab:few_shot_10}
\begin{center}
\resizebox{0.9\textwidth}{!}{%
\begin{tabular}{c|cc|cc|cc|cc|cc|cc|cc}
\toprule

\multirow{2}{*}{Methods}
&\multicolumn{2}{c|}{GPT2(6)-adapter}&\multicolumn{2}{c|}{OFA}&\multicolumn{2}{c|}{DLinear}&\multicolumn{2}{c|}{PatchTST}&\multicolumn{2}{c|}{TimesNet}&\multicolumn{2}{c|}{FEDformer}&\multicolumn{2}{c}{Autoformer} \\
& MSE  & MAE & MSE  & MAE & MSE & MAE& MSE & MAE& MSE  & MAE& MSE  & MAE& MSE  & MAE\\
\midrule

Weather&{\bf0.232}&{\bf0.264}&\underline{0.238} &\underline{0.275} &0.241 &0.283 &0.242 &0.279 &0.279 &0.301 &0.284 &0.324 &0.300 &0.342 \\

ETTh1&{\bf0.576}&{\bf0.429}&\underline{0.590} &\underline{0.525} &0.691 &0.600 &0.633 &0.542 &0.869 &0.628 &0.639 &0.561 &0.702 &0.596 \\

ETTh2&{\bf0.375}&{\bf0.407}&\underline{0.397} &\underline{0.421} &0.605 &0.538 &0.415 &0.431 &0.479 &0.465 &0.466 &0.475 &0.488 &0.499 \\

ETTm1&\underline{0.454}&0.447&0.464 &\underline{0.441} &\textbf{0.411} &\textbf{0.429} &0.501 &0.466 &0.677 &0.537 &0.722 &0.605 &0.802 &0.628 \\

ETTm2&{\bf0.287}&{\bf0.331}&\underline{0.293} &\underline{0.335} &0.316 &0.368 &0.296 &0.343 &0.320 &0.353 &0.463 &0.488 &1.342 &0.930\\


ECL&0.181&\underline{0.273}&\textbf{0.176} &\textbf{0.269} &\underline{0.180} &0.280 &\underline{0.180} &\underline{0.273} &0.323 &0.392 &0.346 &0.427 &0.431 &0.478\\

Traffic&\underline{0.438}&{\bf0.301}&0.440 &0.310 &0.447 &0.313 &\textbf{0.430} &\underline{0.305} &0.951 &0.535 &0.663 &0.425 &0.749 &0.446\\
\midrule

Avg.&
{\bf0.363}&{\bf0.350}& \underline{0.371} &\underline{0.367} &0.413 &0.401 &0.385 &0.376 &0.556 &0.458 &0.511 &0.472 &0.687 &0.559  \\

\bottomrule
\end{tabular}%
}
\vskip -0.20in
\end{center}
\end{table*}

\textbf{Baselines}
We select representative baselines and cite their results from \cite{zhou2023onefitsall}, which includes the most recent and quite extensive empirical studies of time series. The baselines include OFA \cite{zhou2023onefitsall}, TimesNet~\cite{timesnet}, DLinear~\cite{dlinear}, Autoformer~\cite{wu2021autoformer}, FEDformer~\cite{zhou2022fedformer}, ETSformer~\cite{woo2022etsformer}, Non-stationary Transformer~\cite{non-stationary}, Flowformer \cite{huang2022flowformer}, Non-stationary Transformer~\cite{non-stationary}, PatchTST~\cite{Patchformer}. Besides, N-HiTS~\cite{nhits} and N-BEATS~\cite{n-beats} are used for short-term forecasting. Anomaly Transformer~\cite{xu2021anomaly}, DCdetector~\cite{dcdetector}, THOC~\cite{thoc}, InterFusion~\cite{interfusion}, OmniAnomaly~\cite{omnianomaly} and BeatGAN~\cite{beatgan} are used for anomaly detection. Rocket~\cite{ROCKET}, LSTNet~\cite{lstnet} and TCN~\cite{tcn} are used for classification.

\textbf{Main Results}
Overall, with explicit adapters, GPT2-adapter consistently outperforms OFA. This serves as a strong evidence that explicit adapters can also bring significant improvements for time series analysis with LLMs.
Full experiments results can be found in Appendix \ref{app:full_res}


\subsubsection{Time Series Long-term Forecasting}
\textbf{Setups} Eight popular real-world benchmark datasets~\cite{timesnet}, including Weather, Traffic \footnote{http://pems.dot.ca.gov}, Electricity, ILI \footnote{https://gis.cdc.gov/grasp/fluview/fluportaldashboard.html}, and 4 ETT datasets (ETTh1, ETTh2, ETTm1, ETTm2), are used for long-term forecasting evaluation.

\textbf{Results}
As shown in Table \ref{tab:long_term}, GPT2(6)-adapter surpasses all other baselines. Notably, compared with OFA, GPT2(6)-adapter yields a relative \textbf{4.4\%} average MSE reduction.
The results indicates that we successfully leverage both channel-wise/temporal information and frequency information bias through our carefully designed plugin adapter.

\subsubsection{Time Series Short-term Forecasting}

\textbf{Setups}
To fully evaluate different algorithms in forecasting tasks, we also conduct short-term forecasting (with relatively short forecasting horizon) experiments on M4 \cite{makridakis2018m4}, contains marketing data of various frequencies.

\textbf{Results}
The results in Table \ref{tab:short_term} indicate that GPT2(6)-adapter achieves the lowest SMAPE of {\bf 11.713}, outperforming previous SOTA method TimesNet with SMAPE of 11.829.
Also, GPT2(6)-adapter gains an improvement of {\bf 2.3\%} SMAPE relative reduction compared to OFA.
The key difference lies in comparing the performance under long term forecasting and short term forecasting \cite{kim2022reversible,non-stationary}: one with a significant distribution shift and the other without.
In line with empirical results, our proposed method demonstrates strong performance in both scenarios.

\begin{table*}[h]
\caption{Zero-shot Results. Dataset-specific metrics aggregated over each dataset. A lower value indicates better performance. The source dataset of M3, Tourism, Electricity, and M4. For M4, the source data for N-BEATS is FRED, and M3 for other models. \textbf{Black}: best, {\color{red} \textbf{Red}}: second best, {\color{violet} \textbf{Violet}}: third best. Y, Q, M and O are abbreviations for Yearly, Quarterly, Monthly and Others respectively.}
\label{tab:zero_m4}
\begin{center}
\resizebox{1.0\textwidth}{!}{%
\begin{tabular}{c|ccccc|ccccc|cccc|c|c}
\toprule

& \multicolumn{5}{c|}{M4 (sMAPE)} & \multicolumn{5}{c|}{M3 (sMAPE)} & \multicolumn{4}{c|}{TOURISM (MAPE)} & ELECTR  & \multirow{3}{*}{Avg.}\\

& Y & Q & M & O & Avg. & Y & Q & M & O & Avg. & Y & Q & M & Avg. & \multirow{2}{*}{ND$\times$100} \\
& (23k) & (24k) & (48k) & (5k) & (100k) & (645) & (756) & (1428) & (174) & (3003) & (518) & (427) & (366) & (1311) && \\

\midrule

N-BEATS & 13.26 & 9.59 & 12.67 & 4.69 & \textbf{11.67} & 15.07 & 9.07 & 13.19 & 4.29 & \textbf{12.38} & 23.57 & 14.66 & 19.32 & \textbf{18.82} & 17.8 & {\bf15.19}\\
\midrule
DLinear & 14.19 & 18.85 & 14.76 & 9.19 & 15.33 & 17.43 & 9.74 & 15.65 & 6.81 & 14.03 & 39.59 & 18.30 & 24.76 & 28.51 & 17.6 & 18.86\\
TimesNet & 15.65 & 11.87 & 16.16 & 6.86 & 14.55 & 18.75 & 12.26 & 14.01 & 6.88 & 14.17 & 35.59 & 19.22 & 30.54 & 28.84 &19.3 & 18.96\\
PatchTST & 13.96 & 10.92 & 14.66 & 7.08 & 13.22 &  15.99 & 9.62 & 14.71 & 9.44 & 13.39 & 33.23 & 19.27 & 27.57 & 27.10 &\color{violet} \textbf{17.3} & 17.67\\
FEDformer & 13.88 & 11.51 & 18.15 & 7.52 & 15.04 & 16.00 & 9.48 & 15.12 & 8.94 & 13.53 & 43.41 & 19.88 & 28.39 & 31.55 &18.4 & 19.63\\
Autoformer & 14.55 & 17.34 & 25.06 & 9.66 & 20.02 & 16.18 & 13.92 & 16.91 & 14.68 & 15.87 & 51.19 & 34.95 & 31.47 & 40.39 &33.9 & 27.54\\
OFA&13.74&10.78&14.63&7.08&{\color{violet} \textbf{13.12}}&16.42&10.13&14.10&4.81&{\color{violet} \textbf{13.06}}&27.17&16.21&21.92&{\color{red} \textbf{22.14}}&\color{red} \textbf{17.2}&\color{violet}\textbf{16.38} \\
\midrule
GPT2(6)- & \multirow{2}{*}{13.64} & \multirow{2}{*}{10.65} & \multirow{2}{*}{14.64} & \multirow{2}{*}{6.99} & \multirow{2}{*}{{\color{red} \textbf{13.07}}} & \multirow{2}{*}{15.74} & \multirow{2}{*}{9.53} & \multirow{2}{*}{13.95} & \multirow{2}{*}{4.76} & \multirow{2}{*}{{\color{red} \textbf{12.52}}} & \multirow{2}{*}{28.58} & \multirow{2}{*}{15.58} & \multirow{2}{*}{21.95} & \multirow{2}{*}{{\color{violet} \textbf{22.49}}}&\multirow{2}{*}{{\bf16.3}}&\multirow{2}{*}{\color{red}\textbf{16.09}} \\
-adapter &&&&&&&&&&&&&&&\\

\bottomrule
\end{tabular}%
}
\vskip 0in
\end{center}
\end{table*}

\subsubsection{Time Series Anomaly Detection}

\textbf{Setups}
We compare models on five commonly used datasets, including SMD\cite{SMD}, MSL\cite{MSL_SMAP}, SMAP\cite{MSL_SMAP}, SWaT\cite{SWaT} and PSM\cite{PSM}. 

\textbf{Results}
Table \ref{tab:anomaly_sota} demonstrates that GPT2(6)-adapter achieves the best performance with the averaged F1-score \textbf{95.35\%}, surpassing previous SOTA time series anomaly detection methods DCdetector~\cite{dcdetector} (95.01\%) and Anomaly Transformer~\cite{xu2021anomaly} (94.91\%).
Among baselines, OFA and TimesNet utilize reconstruction error only for simplicity, while DCdetecor and Anomaly Transoformer introduce association discrepancy.
Therefore, with a remarkable {\bf 8.63\%} improvement compared to OFA, it can be concluded that besides forecasting, explicit adapters can also be effective in anomaly detection.

\subsubsection{Time Series Classification}

\textbf{Setups} To evaluate the model's capacity for high-level representation learning, we employ sequence-level classification. Specifically, we follow the same setting as TimesNet: For classification, 10 multivariate UEA classification datasets \cite{UEA} are selected for evaluation, including gesture, action, audio recognition medical diagnosis and other practical tasks.

\textbf{Results} As shown in Table \ref{tab:classification}, GPT2(6)-adapter achieves an average accuracy of {\bf 74.1\%}, surpassing all baselines including OFA (74.0\%) and TimesNet (73.60\%), which shows the explicit adapters can indeed help in high-level representation.

\subsubsection{Time Series Imputation}

\textbf{Setups} 
We conduct experiments on six popular real-world datasets, including 4 ETT datasets \cite{zhou2021informer} (ETTh1, ETTh2, ETTm1, ETTm2), Electricity and Weather, where the data-missing is common. Following the settings of TimesNet, different random mask ratios (\{12.5\%, 25\%, 37.5\%, 50\%\}) of time points are selected for the evaluation on various proportions of missing data.

\textbf{Results} 
The results are shown in Table \ref{tab:imputation} that GPT2(3)-adapter both outperforms the previous methods. Particularly, compared to the previous SOTA OFA, GPT2(3)-adapter yields a relative \textbf{11.1\%} MSE reduction on ECL,and a \textbf{6.2\%} MSE reduction on average on six benchmark datasets.

\subsubsection{Few-shot Forecasting}
\textbf{Setups} 
The large language model (LLM) has demonstrated remarkable performance in both few-shot and zero-shot learning settings ~\cite{Brown2020LanguageMA,OpenAI2023GPT4TR}. Thus, for the few-shot forecasting, we follow the experimental setting of OFA and use 10\% percentage time step for comparison. 

\textbf{Results}
The results are shown in Table \ref{tab:few_shot_10}. Compared to OFA, TimesNet, PatchTST and other methods, GPT2(6)-adapter achieves better performance. Traditionally, CNN-based and single MLP-based models are considered more data-efficient for training and suitable for few-shot learning methods. In comparison to convolution-based TimesNet, MLP-based DLinear and transformer-based OFA, GPT2(6)-adapter demonstrates relative average MSE reductions of \textbf{34.7\%}, \textbf{12.1\%} and \textbf{2.1\%} respectively. 

\subsubsection{Zero-shot Forecasting} 

\textbf{Setups} 
This task is used to evaluate the cross datasets adaption ability of our proposed algorithm, i.e. how well a model is able to perform on dataset $A$ (without any training data from $A$) when it is trained from dataset $B$. 

\textbf{Results}
The results are summarized in Table \ref{tab:zero_m4}. GPT2(6)-adapter model consistently outperforms all recent state-of-the-art transformer and MLP-based time series forecasting methods. Compared to TimesNet, PatchTST and OFA, GPT2(6)-adapter demonstrates a relative average metric reduction of \textbf{15.1\%}, \textbf{8.9\%} and \textbf{1.8\%}, respectively. 
Also, GPT2(6)-adapter is comparable to N-BEATS without any meta-learning design and outperform N-BEATS in the ELECTR dataset.
Zero-shot learning is undoubtedly a promising direction that warrants exploration. The concept of learning without re-training the model, known as zero-shot learning or in-context learning, has been presented to us by LLM (Language Model). We firmly believe that the time series domain is no exception and holds significant potential for exploration. Our proposed method represents a small but meaningful step towards this path.

\subsubsection{Ablation Experiments}

Due to space constraints, we have consolidated the ablation experiments for both datasets into Table \ref{tab:ablation_adapter} to demonstrate the effectiveness of the designed adapters. Each of these adapters is effective in enhancing performance, and the introduction of all adapters yields the most substantial improvement.

\begin{table}
\caption{Ablation on adapters.}
\vskip -0.10in
\label{tab:ablation_adapter}
\begin{center}
\resizebox{0.40\textwidth}{!}{%
\begin{tabular}{c|ccc|cc}
\hline

\multirow{2}{*}{Datasets}
& \multicolumn{3}{c|}{Adapters} & \multirow{2}{*}{MSE} & \multirow{2}{*}{MAE}\\
& Temporal & Channel & Frequency \\

\toprule

\multirow{5}{*}{ETTh1 96}
& - & - & - & 0.376 & 0.397 \\
& \checkmark & - & - & 0.375 & 0.400 \\
& - & \checkmark & - & 0.373 & 0.398 \\
& - & - & \checkmark & 0.369 & 0.397 \\
& \checkmark & \checkmark & \checkmark & {\bf0.366} & {\bf0.394} \\

\midrule

\multirow{5}{*}{ETTh2 96}
& - & - & - & 0.285 & 0.342 \\
& \checkmark & - & - & 0.280 & 0.341\\
& - & \checkmark & - & 0.285 & 0.343 \\
& - & - & \checkmark & 0.282 & 0.344\\
& \checkmark & \checkmark & \checkmark & {\bf0.269} & {\bf0.331} \\

\bottomrule

\end{tabular}
}
\vskip -0.3in
\end{center}
\end{table} 
\section{Conclusions}\label{sec:conclusions}
In this study, we investigated the role of exit methods using textual prompts to enhance time series forecasting with large language models (LLMs). Our findings suggest that the true value of these prompts may lie in their ability to introduce new parameters that act as implicit adapters, effectively bridging the domain gap. We propose two random testing methods to evaluate the effectiveness of textual information. Additionally, we leverage the inductive biases inherent in time series data and tasks to develop four customized adapters, which significantly improve prediction accuracy and outperform traditional methods. This marks a dual contribution: elucidating how textual prompts facilitate LLM forecasting and establishing the efficacy of adapter-based approaches. Our work clarifies the role of LLMs in time series analysis and prompts further exploration into adapter innovation and LLM applications.

\bibliography{main}
\bibliographystyle{plain}

\newpage
\appendix
\onecolumn
\section{Appendix}

\subsection{Experiments}
\label{app:exp}

\subsubsection{LLaMA(32) with Random Prompting}
\label{app:llama32}
As shown in Table \ref{tab:llama32}, for LLaMA(32) with random prompting, even with only 4-bit precision, it outperforms the standard prompts by \textbf{3.2}\%
\begin{table}[h!]
\caption{Performance (MSE) with random prompting. S and RP represent Standard and Random Prompting respectively.}
\vskip -0.10in
\label{tab:llama32}
\begin{center}
\scalebox{0.8}{
\begin{tabular}{c|ccc}

\toprule








& \multicolumn{3}{c}{LLaMA(32)} \\

& S&S(w/o Dataset Context)&RP (4bit) \\

\midrule

ETTh1 96&0.362&0.402&0.389 \\
ETTh1 192 &0.398&0.417&0.414 \\
\midrule
ETTm1 96 &0.272&0.298&0.311 \\
ETTm1 192 &0.310&0.331&0.350 \\

\bottomrule

\end{tabular}
}
\end{center}
\vskip -0.10in
\end{table}

\subsubsection{Mean and STD for Few-shot Learning}
Table \ref{tab:std_results} lists both mean and STD for GPT2(6)-adapter on ETTh1 96 and 192. The results show a small variance in performance of GPT2(6)-adapter that represents the stability of GPT2(6)-adapter.

\begin{table}[h]
\caption{A subset of results showing both Mean and STD on ETTh1 96 and ETTh1 192.}
\label{tab:std_results}
\vskip 0.15in
\begin{center}
\begin{small}
\scalebox{1}{
\begin{tabular}{c|ccc}
\toprule

Methods&\multicolumn{2}{c}{GPT2(6)-adapter} \\ 


Metric & MSE  & MAE \\ 
\midrule

ETTh1 96 &0.367$\pm$0.0012&0.394$\pm$0.0004\\
ETTh1 192 &0.406$\pm$0.0008&0.418$\pm$0.0002 \\

\bottomrule
\end{tabular}
}
\end{small}
\end{center}
\vskip -0.1in
\end{table}

\subsection{Full Results}
\label{app:full_res}
\begin{table}[ht]
\caption{Anomaly detection task. We calculate the F1-score, Precision and Recall (as \%) for each dataset. A higher F1 indicates better performance. \textbf{Black}: best, \underline{Underline}: second best.}
\label{tab:anomaly_sota_full}
\begin{center}
\resizebox{\textwidth}{!}{%
\begin{tabular}{cc|ccccccccc}
\toprule

\multicolumn{2}{c|}{{Methods}}&GPT2(6)-adapter&OFA&DCdetector&Anomaly&TimeNet&THOC&InterFusion&OmniAnomaly&BeatGAN \\
\midrule
\multirow{3}{*}{\rotatebox{90}{$SMD$}}
&P&88.65&88.89&83.59&89.40&87.91&79.76&87.02&83.68&72.90 \\
&R&91.37&84.98&91.10&95.45&82.54&90.95&85.43&86.82&84.09 \\
&F1&\underline{89.99}&86.89&87.18&\textbf{92.33}&84.61&84.99&86.22&85.22&78.10 \\
\midrule

\multirow{3}{*}{\rotatebox{90}{$MSL$}}
&P&93.26&82.00&93.69&92.09&89.54&88.45&81.28&89.02&89.75 \\
&R&93.90&82.91&99.69&95.15&75.36&90.97&92.70&86.37&85.42 \\
&F1&\underline{93.60}&82.45&\textbf{96.60}&93.59&81.84&89.69&86.62&87.67&87.53 \\
\midrule

\multirow{3}{*}{\rotatebox{90}{$SMAP$}}
&P&95.43&90.60&95.63&94.13&90.14&92.06&89.77&92.49&92.38 \\
&R&98.38&60.95&98.92&99.40&56.40&89.34&88.52&81.99&55.85 \\
&F1&\underline{96.88}&72.88&\textbf{97.02}&96.69&69.39&90.68&89.14&86.92&69.60 \\
\midrule

\multirow{3}{*}{\rotatebox{90}{$SWaT$}}
&P&96.44&92.20&93.11&91.55&90.75&83.94&80.59&81.42&64.01 \\
&R&99.78&96.34&99.77&96.73&95.40&86.36&85.58&84.30&87.76 \\
&F1&\textbf{98.08}&94.23&\underline{96.33}&94.07&93.02&85.13&83.01&82.83&79.92 \\
\midrule

\multirow{3}{*}{\rotatebox{90}{$PSM$}}
&P&99.06&98.62&97.14&96.91&98.51&88.14&83.61&88.39&90.30 \\
&R&97.39&95.68&98.74&98.90&96.20&90.99&83.45&74.46&93.84 \\
&F1&\textbf{98.22}&97.13&\underline{97.94}&97.89&97.34&89.54&83.52&80.83&92.83 \\
\midrule

\multicolumn{2}{c|}{$Avg.\ F1$}&\textbf{95.35}&86.72&\underline{95.01} &94.91&85.24&88.01&85.70&84.69&81.60 \\
\midrule
\end{tabular}%
}
%

\end{center}
\end{table}
\begin{table*}[ht]
\caption{Full results for the classification task. $\ast$. in the Transformers indicates the name of $\ast$former. A higher accuracy indicates better performance. \textbf{Black}: best, \underline{Underline}: second best.}
\label{tab:classification_full}
\begin{center}
\resizebox{\textwidth}{!}{%
\setlength\tabcolsep{3pt}
\begin{tabular}{c|c|c|c|ccccc|c|c|cc}
\toprule

\multirow{2}{*}{Methods} &
Classical & RNN & \multirow{2}{*}{TCN}& \multicolumn{5}{c|}{Transformers}  & \multirow{2}{*}{DLinear}&\multirow{2}{*}{TimesNet} & \multirow{2}{*}{OFA}& \multirow{2}{*}{GPT2(6)-adapter} \\
&Rocket&LSTNet& & Auto.& Station.& FED.& ETS.& Flow. & & &  \\

\midrule
EthanolConcentration &45.2 &39.9 &28.9 &31.6 &32.7 &31.2 &28.1 &33.8 &32.6 &\textbf{35.7} &34.2&\underline{35.0}\\
FaceDetection &64.7 &65.7 &52.8 &68.4 &68.0 &66.0 &66.3 &67.6 &68.0 &68.6 &\underline{69.2}&\textbf{69.7}\\
Handwriting &58.8 &25.8 &53.3 &36.7 &31.6 &28.0 &32.5 &\textbf{33.8} &27.0 &32.1 &\underline{32.7}&32.2\\
Heartbeat &75.6 &77.1 &75.6 &74.6 &73.7 &73.7 &71.2 &77.6 &75.1 &\underline{78.0} &77.2&\textbf{79.5}\\
JapaneseVowels &96.2 &98.1 &98.9 &96.2 &99.2 &98.4 &95.9 &\textbf{98.9} &96.2 &98.4 &\underline{98.6}&98.1\\
PEMS-SF &75.1 &86.7 &86.1 &82.7 &87.3 &80.9 &86.0 &83.8 &75.1 &\textbf{89.6} &\underline{87.9}&86.1\\
SelfRegulationSCP1 &90.8 &84.0 &84.6 &84.0 &89.4 &88.7 &89.6 &92.5 &87.3 &91.8 &\textbf{93.2}&\textbf{93.2}\\
SelfRegulationSCP2 &53.3 &52.8 &55.6 &50.6 &57.2 &54.4 &55.0 &56.1 &50.5 &57.2 &\underline{59.4}&\textbf{60.6}\\
SpokenArabicDigits &71.2 &\textbf{100.0} &95.6 &\textbf{100.0} &\textbf{100.0} &\textbf{100.0} &\textbf{100.0} &98.8 &81.4 &99.0 &99.2&98.7\\
UWaveGestureLibrary &\textbf{94.4} &87.8 &\underline{88.4} &85.9 &87.5 &85.3 &85.0 &86.6 &82.1 &85.3 &88.1&87.5\\
\midrule
Average &72.5 &71.8 &70.3 &71.1 &72.7 &70.7 &71.0 &73.0 &67.5 &73.6&\underline{74.0}&\textbf{74.1}\\
\bottomrule

\end{tabular}%
}
\end{center}
\end{table*}
\begin{table*}[ht]
\renewcommand\arraystretch{1.5}
\caption{Short-term forecasting task on M4. The prediction lengths are in [6, 48] and results are weighted averaged from several datasets under different sample intervals. A lower SMAPE value indicates better performance. \textbf{Black}: best, \underline{Underline}: second best.}
\label{tab:short_term_full}
\begin{center}
\resizebox{\textwidth}{!}{%
\begin{tabular}{cc|cccccccccc}
\toprule

\multicolumn{2}{c|}{Methods}&GPT2(6)-adapter&GPT2(6)-frozen&TimesNet&PatchTST&N-HiTS&N-BEATS& DLinear &FEDformer &Autoformer \\


\midrule
\multirow{3}{*}{\rotatebox{90}{$Yearly$}}
&SMAPE&{\bf13.288}&13.531&\underline{13.387}&13.477&13.418&13.436&16.965&13.728&13.974\\
&MASE&\underline{3.005}&3.015&{\bf2.996}&3.019&3.045&3.043&4.283&3.048&3.134\\
&OWA&{\bf0.785}&0.793&\underline{0.786}&0.792&0.793&0.794&1.058&0.803&0.822\\
\bottomrule

\multirow{3}{*}{\rotatebox{90}{$Quarterly$}}
&SMAPE&{\bf9.955}&10.177&\underline{10.100}&10.380&10.202&10.124&12.145&10.792&11.338\\
&MASE&{\bf1.162}&1.194&\underline{1.182}&1.233&1.194&1.169&1.520&1.283&1.365\\
&OWA&{\bf0.876}&0.898&0.890&0.921&0.899&\underline{0.886}&1.106&0.958&1.012\\
\bottomrule

\multirow{3}{*}{\rotatebox{90}{$Monthly$}}
&SMAPE&{\bf12.599}&12.894&\underline{12.670}&12.959&12.791&12.677&13.514&14.260&13.958\\
&MASE&{\bf0.933}&0.956&{\bf0.933}&0.970&0.969&1.053&1.037&1.102&1.103\\
&OWA&{\bf0.876}&0.897&\underline{0.878}&0.905&0.899&0.880&0.956&1.012&1.002\\
\bottomrule

\multirow{3}{*}{\rotatebox{90}{$Others$}}
&SMAPE&{\bf4.420}&4.940&\underline{4.891}&4.952&5.061&4.925&6.709&4.954&5.485\\
&MASE&{\bf3.101}&3.228&3.302&3.347&\underline{3.216}&3.391&4.953&3.264&3.865\\
&OWA&{\bf0.954}&\underline{1.029}&1.035&1.049&1.040&1.053&1.487&1.036&1.187\\
\bottomrule

\multirow{3}{*}{\rotatebox{90}{$Average$}}
&SMAPE &{\bf11.713}&11.991 &\underline{11.829}&12.059& 11.927& 11.851& 13.639 &12.840 &12.909  \\
&MASE &{\bf1.572}& 1.600 & \underline{1.585}&1.623 & 1.613 & 1.599  &2.095 &1.701 &1.771  \\
&OWA &{\bf0.858}&0.879 & \underline{0.867}&0.890 &0.881 &0.870  &1.051 &0.952 &0.972 \\

\bottomrule

\end{tabular}%
}
\end{center}
\end{table*}
\begin{table*}[ht]
\caption{Long-term forecasting task. We use prediction length $O \in \{96, 192, 336, 720\}$. A lower MSE indicates better performance. \textbf{Black}: best, \underline{Underline}: second best.}
\label{tab:long_term_full}
\begin{center}
\resizebox{\textwidth}{!}{%

\setlength\tabcolsep{3pt}
\begin{tabular}{c|c|cc|cc|cc|cc|cc|cc|cc|cc}
\toprule

\multicolumn{2}{c|}{Methods}&\multicolumn{2}{c|}{GPT2(6)-adapter}&\multicolumn{2}{c|}{OFA}&\multicolumn{2}{c|}{GPT2(0)}&\multicolumn{2}{c|}{DLinear}&\multicolumn{2}{c|}{PatchTST}&\multicolumn{2}{c|}{TimesNet}&\multicolumn{2}{c|}{FEDformer}&\multicolumn{2}{c}{Autoformer} \\

\midrule

\multicolumn{2}{c|}{Metric} & MSE  & MAE & MSE & MAE& MSE & MAE& MSE  & MAE& MSE  & MAE& MSE  & MAE& MSE  & MAE& MSE  & MAE \\
\midrule

\multirow{5}{*}{\rotatebox{90}{$Weather$}}
& 96  &0.144&0.183& 0.162 & 0.212 & 0.181 & 0.232 & 0.176 & 0.237 & 0.149 & 0.198 &0.172&0.220&0.217&0.296&0.266&0.336\\
& 192 &0.188&0.228& 0.204 & 0.248 & 0.222 & 0.266 & 0.220 & 0.282 & 0.194 & 0.241 &0.219&0.261&0.276&0.336&0.307&0.367\\
& 336 &0.239&0.268& 0.254 & 0.286 & 0.270 & 0.299 & 0.265 & 0.319 & 0.245 & 0.282&0.280&0.306&0.339&0.380&0.359&0.395 \\
& 720 &0.308&0.321& 0.326 & 0.337 & 0.338 & 0.345 & 0.333 & 0.362 & 0.314 & 0.334&0.365&0.359&0.403&0.428&0.419&0.428\\
& Avg &{\bf0.219}&{\bf0.250}&0.237&0.270&0.252&0.285&0.248&0.300&\underline{0.225}&\underline{0.264}&0.259&0.287&0.309&0.360&0.338&0.382\\
\midrule

\multirow{5}{*}{\rotatebox{90}{$ETTh1$}}
& 96  &0.367&0.394& 0.376 & 0.397 & 0.422 & 0.428 & 0.375 & 0.399 & 0.370 & 0.399 &0.384&0.402&0.376&0.419&0.449&0.459\\
& 192 &0.406&0.418& 0.416 & 0.418 & 0.466 & 0.450 & 0.405 & 0.416 & 0.413 & 0.421&0.436&0.429&0.420&0.448&0.500&0.482\\
& 336 &0.420&0.439& 0.442 & 0.433 & 0.488 & 0.464 &0.439 & 0.443 & 0.422 & 0.436 &0.491&0.469&0.459&0.465&0.521&0.496\\
& 720 &0.432&0.455& 0.477 & 0.456 & 0.485 & 0.478 & 0.472 & 0.490 & 0.447 & 0.466 &0.521&0.500&0.506&0.507&0.514&0.512\\
& Avg &{\bf0.406}&\underline{0.427}&0.427&\textbf{0.426}&0.465&0.455&0.422&0.437&\underline{0.413}&0.430&0.458&0.450&0.440&0.460&0.496&0.487\\
\midrule

\multirow{5}{*}{\rotatebox{90}{$ETTh2$}}
& 96  &0.269&0.331& 0.285 & 0.342 & 0.318 & 0.368 & 0.289 & 0.353 & 0.274 & 0.336&0.340&0.374&0.358&0.397&0.346&0.388 \\
& 192 &0.334&0.379& 0.354 & 0.389 & 0.383 & 0.407 & 0.383 & 0.418 & 0.339 &  0.379 &0.402&0.414&0.429&0.439&0.456&0.452\\
& 336 &0.359&0.398& 0.373 & 0.407 & 0.406 & 0.427 & 0.448 & 0.465 & 0.329 & 0.380&0.452&0.452&0.496&0.487&0.482&0.486 \\
& 720 &0.392&0.433& 0.406 & 0.441 & 0.420 & 0.446 & 0.605 & 0.551 & 0.379 & 0.422&0.462&0.468&0.463&0.474&0.515&0.511\\
& Avg &\underline{0.338}&\underline{0.385}&0.354&0.394&0.381&0.412&0.431&0.446&\textbf{0.330}&\textbf{0.379}&0.414&0.427&0.437&0.449&0.450&0.459\\
\midrule

\multirow{5}{*}{\rotatebox{90}{$ETTm1$}}
& 96  &0.292&0.339& 0.292 & 0.346 & 0.330 & 0.372 & 0.299 & 0.343 & 0.290 & 0.342 &0.338&0.375&0.379&0.419&0.505&0.475 \\
& 192 &0.330&0.363& 0.332 & 0.372 & 0.371 & 0.394 & 0.335 & 0.365 & 0.332 & 0.369&0.374&0.387&0.426&0.441&0.553&0.496\\
& 336 &0.360&0.379& 0.366 & 0.394 & 0.398 & 0.409 & 0.369 & 0.386 & 0.366 & 0.392&0.410&0.411&0.445&0.459&0.621&0.537\\
& 720 &0.413&0.406& 0.417 & 0.421 & 0.454 & 0.440 & 0.425 & 0.421 & 0.416 & 0.420&0.478&0.450&0.543&0.490&0.671&0.561\\
&Avg&{\bf0.348}&{\bf0.371}&0.352&0.383&0.388&0.403&0.357&\underline{0.378}&\underline{0.351}&0.380&0.400&0.406&0.448&0.452&0.588&0.517 \\
\midrule

\multirow{5}{*}{\rotatebox{90}{$ETTm2$}}
& 96  &0.160&0.247& 0.173 & 0.262 & 0.192 & 0.281 & 0.167 & 0.269 & 0.165 & 0.255&0.187&0.267&0.203&0.287&0.255&0.339\\
& 192 &0.212&0.287& 0.229 & 0.301 & 0.245 & 0.317 & 0.224 & 0.303 & 0.220 & 0.292&0.249&0.309&0.269&0.328&0.281&0.340\\
& 336 &0.264&0.319& 0.286 & 0.341 & 0.302 & 0.352 & 0.281 & 0.342 & 0.274 & 0.329 &0.321&0.351&0.325&0.366&0.339&0.372\\
& 720 &0.355&0.376& 0.378 & 0.401 & 0.399 & 0.408 & 0.397 & 0.421 & 0.362 & 0.385&0.408&0.403&0.421&0.415&0.433&0.432 \\
& Avg&{\bf0.247}&{\bf0.307}&0.266&0.326&0.284&0.339&0.267&0.333&\underline{0.255}&\underline{0.315}&0.291&0.333&0.305&0.349&0.327&0.371 \\
\midrule


\multirow{5}{*}{\rotatebox{90}{$ECL$}}
& 96  &0.131&0.225& 0.139 & 0.238 &0.138 & 0.234 & 0.140 & 0.237 & 0.129 & 0.222&0.168&0.272&0.193&0.308&0.201&0.317 \\
& 192 &0.151&0.245& 0.153 & 0.251 & 0.152 & 0.247 & 0.153 & 0.249 & 0.157 &0.240&0.184&0.289&0.201&0.315&0.222&0.334\\
& 336 &0.162&0.254& 0.169 & 0.266 & 0.168 & 0.263 & 0.169 & 0.267 & 0.163 & 0.259&0.198&0.300&0.214&0.329&0.231&0.338\\
& 720 &0.192&0.284& 0.206 & 0.297 & 0.207 & 0.295 & 0.203 & 0.301 & 0.197 & 0.290&0.220&0.320&0.246&0.355&0.254&0.361\\
& Avg &\textbf{0.159}&\textbf{0.252}&0.167&0.263&0.166&0.259&0.166&0.263&\underline{0.161}&\textbf{0.252}&0.192&0.295&0.214&0.327&0.227&0.338\\
\midrule

\multirow{5}{*}{\rotatebox{90}{$Traffic$}}
& 96  &0.378&0.250& 0.388 & 0.282 & 0.390 & 0.272 & 0.410 & 0.282 & 0.360 & 0.249&0.593&0.321&0.587&0.366&0.613&0.388\\
& 192 &0.384&0.248& 0.407 & 0.290 &0.403 & 0.276 & 0.423 & 0.287 &0.379 & 0.256&0.617&0.336&0.604&0.373&0.616&0.382\\
& 336 &0.393&0.255& 0.412 & 0.294 & 0.413 & 0.280 & 0.436 & 0.296 & 0.392 & 0.264&0.629&0.336&0.621&0.383&0.622&0.337 \\
& 720 &0.434&0.276& 0.450 & 0.312 & 0.447 & 0.298 & 0.466 & 0.315 & 0.432 & 0.286&0.640&0.350&0.626&0.382&0.660&0.408\\
& Avg &\underline{0.397}&{\bf0.257}&0.414&0.294&0.413&0.281&0.433&0.295&\textbf{0.390}&\underline{0.263}&0.620&0.336&0.610&0.376&0.628&0.379\\
\midrule
\multicolumn{2}{c|}{Average}&{\bf0.302}&{\bf0.321}&0.316&0.336&0.335&0.347&0.332&0.350&\underline{0.304}&\underline{0.326}&0.376&0.362&0.394&0.396&0.436&0.419\\

\bottomrule
\end{tabular}%
}

\end{center}
\end{table*}
\begin{table*}[ht]
\caption{ Imputation task. We randomly mask {12.5\%, 25\%, 37.5\%, 50\%} time points of 96-length time series. 
A lower MSE indicates better performance. \textbf{Black}: best, \underline{Underline}: second best.}
\label{tab:imputation_full}
\begin{center}
\resizebox{\textwidth}{!}{%
\setlength\tabcolsep{3pt}
\begin{tabular}{cc|cc|cc|cc|cc|cc|cc|cc|cc}
\toprule

\multicolumn{2}{c|}{Methods} &\multicolumn{2}{c|}{GPT2(3)-adapter}
&\multicolumn{2}{c|}{GPT2(3)-frozen} & \multicolumn{2}{c|}{TimesNet}& \multicolumn{2}{c|}{PatchTST}&\multicolumn{2}{c|}{DLinear}&\multicolumn{2}{c|}{FEDformer}&\multicolumn{2}{c|}{Stationary}&\multicolumn{2}{c}{Autoformer} \\
Mask&Ratio&MSE&MAE&MSE&MAE&MSE&MAE&MSE&MAE&MSE&MAE&MSE&MAE&MSE&MAE&MSE&MAE \\

\midrule
\multirow{5}{*}{\rotatebox{90}{$ETTm1$}}
& 12.5\% &0.018&0.091& 0.017&0.085&0.023&0.101&0.041&0.130&0.080&0.193&0.052&0.166&0.032&0.119&0.046&0.144 \\
& 25\% &0.022&0.099&0.022&0.096&0.023&0.101&0.044&0.135&0.080&0.193&0.052&0.166&0.032&0.119&0.046&0.144 \\
& 37.5\% &0.026&0.107&0.029&0.111&0.029&0.111&0.049&0.143&0.103&0.219&0.069&0.191&0.039&0.131&0.057&0.161 \\
& 50\% &0.034&0.123&0.040&0.128&0.036&0.124&0.055&0.151&0.132&0.248&0.089&0.218&0.047&0.145&0.067&0.174 \\
& Avg &{\bf0.025}&{\bf0.105}&\underline{0.028}&{\bf0.105}&0.027&0.107&0.047&0.140&0.093&0.206&0.062&0.177&0.036&0.126&0.051&0.150 \\
\midrule

\multirow{5}{*}{\rotatebox{90}{$ETTm2$}}
& 12.5\% &0.019&0.081&0.017&0.076&0.018&0.080&0.026&0.094&0.062&0.166&0.056&0.159&0.021&0.088&0.023&0.092 \\
& 25\% &0.021&0.088&0.020&0.080&0.020&0.085&0.028&0.099&0.085&0.196&0.080&0.195&0.024&0.096&0.026&0.101 \\
& 37.5\% &0.024&0.094&0.022&0.087&0.023&0.091&0.030&0.104&0.106&0.222&0.110&0.231&0.027&0.103&0.030&0.108 \\
& 50\% &0.027&0.102&0.025&0.095&0.026&0.098&0.034&0.110&0.131&0.247&0.156&0.276&0.030&0.108&0.035&0.119 \\
& Avg &\underline{0.022}&0.091&{\bf0.021}&{\bf0.084}&\underline{0.022}&\underline{0.088}&0.029&0.102&0.096&0.208&0.101&0.215&0.026&0.099&0.029&0.105 \\
\midrule

\multirow{5}{*}{\rotatebox{90}{$ETTh1$}}
& 12.5\% &0.040&0.137&0.043&0.140&0.057&0.159&0.093&0.201&0.151&0.267&0.070&0.190&0.060&0.165&0.074&0.182 \\
& 25\% &0.056&0.161&0.054&0.156&0.069&0.178&0.107&0.217&0.180&0.292&0.106&0.236&0.080&0.189&0.090&0.203 \\
& 37.5\% &0.072&0.182&0.072&0.180&0.084&0.196&0.120&0.230&0.215&0.318&0.124&0.258&0.102&0.212&0.109&0.222 \\
& 50\% &0.105&0.219&0.107&0.216&0.102&0.215&0.141&0.248&0.257&0.347&0.165&0.299&0.133&0.240&0.137&0.248 \\
& Avg &{\bf0.068}&\underline{0.174}&\underline{0.069}&{\bf0.173}&0.078&0.187&0.115&0.224&0.201&0.306&0.117&0.246&0.094&0.201&0.103&0.214 \\
\midrule

\multirow{5}{*}{\rotatebox{90}{$ETTh2$}}
& 12.5\% &0.027&0.122&0.039&0.125&0.040&0.130&0.057&0.152&0.100&0.216&0.095&0.212&0.042&0.133&0.044&0.138 \\
& 25\% &0.041&0.130&0.044&0.135&0.046&0.141&0.061&0.158&0.127&0.247&0.137&0.258&0.049&0.147&0.050&0.149 \\
& 37.5\% &0.047&0.141&0.051&0.147&0.052&0.151&0.067&0.166&0.158&0.276&0.187&0.304&0.056&0.158&0.060&0.163 \\
& 50\% &0.054&0.152&0.059&0.158&0.060&0.162&0.073&0.174&0.183&0.299&0.232&0.341&0.065&0.170&0.068&0.173 \\
& Avg &{\bf0.045}&{\bf0.136}&\underline{0.048}&\underline{0.141}&0.049&0.146&0.065&0.163&0.142&0.259&0.163&0.279&0.053&0.152&0.055&0.156 \\
\midrule

\multirow{5}{*}{\rotatebox{90}{$ECL$}}
& 12.5\% &0.066&0.178&0.080&0.194&0.085&0.202&0.055&0.160&0.092&0.214&0.107&0.237&0.093&0.210&0.089&0.210 \\
& 25\% &0.075&0.191&0.087&0.203&0.089&0.206&0.065&0.175&0.118&0.247&0.120&0.251&0.097&0.214&0.096&0.220 \\
& 37.5\%  &0.085&0.203&0.094&0.211&0.094&0.213&0.076&0.189&0.144&0.276&0.136&0.266&0.102&0.220&0.104&0.229\\
& 50\% &0.933&0.212&0.101&0.220&0.100&0.221&0.091&0.208&0.175&0.305&0.158&0.284&0.108&0.228&0.113&0.239 \\
& Avg &{\bf0.080}&{\bf0.196}&\underline{0.090}&\underline{0.207}&0.092&0.210&0.072&0.183&0.132&0.260&0.130&0.259&0.100&0.218&0.101&0.225 \\
\midrule

\multirow{5}{*}{\rotatebox{90}{$Weather$}}
& 12.5\% &0.026&0.046&0.026&0.049&0.025&0.045&0.029&0.049&0.039&0.084&0.041&0.107&0.027&0.051&0.026&0.047 \\
& 25\% &0.029&0.052&0.028&0.052&0.029&0.052&0.031&0.053&0.048&0.103&0.064&0.163&0.029&0.056&0.030&0.054 \\
& 37.5\% &0.031&0.057&0.033&0.060&0.031&0.057&0.035&0.058&0.057&0.117&0.107&0.229&0.033&0.062&0.032&0.060 \\
& 50\% &0.035&0.064&0.037&0.065&0.034&0.062&0.038&0.063&0.066&0.134&0.183&0.312&0.037&0.068&0.037&0.067 \\
& Avg &{\bf0.030}&\underline{0.055}&0.031&0.056&{\bf0.030}&{\bf0.054}&0.060&0.144&0.052&0.110&0.099&0.203&0.032&0.059&0.031&0.057 \\
\midrule

\multicolumn{2}{c|}{Average}&{\bf0.045}&{\bf0.126}&\underline{0.048}&\underline{0.128}&0.050&0.132&0.064&0.159&0.119&0.224&0.112&0.229&0.056&0.142&0.061&0.151 \\
\bottomrule

\end{tabular}%
}
\end{center}
\end{table*}
\begin{table*}[!ht]
\caption{Few-shot learning results on 10\% data. We use prediction length $O \in \{96, 192, 336, 720\}$. A lower MSE indicates better performance. \textbf{Black}: best, \underline{Underline}: second best.}
\setlength\tabcolsep{3pt}
\label{tab:few_shot_10_full}
\begin{center}
\resizebox{\textwidth}{!}{%
\begin{tabular}{c|c|cc|cc|cc|cc|cc|cc|cc|cc}
\toprule

\multicolumn{2}{c|}{Methods}&\multicolumn{2}{c|}{GPT2(6)-adapter}&\multicolumn{2}{c|}{GPT2(6)-frozen}&\multicolumn{2}{c|}{GPT2(0)}&\multicolumn{2}{c|}{DLinear}&\multicolumn{2}{c|}{PatchTST}&\multicolumn{2}{c|}{TimesNet}&\multicolumn{2}{c|}{FEDformer}&\multicolumn{2}{c}{Autoformer} \\

\midrule

\multicolumn{2}{c|}{Metric} & MSE  & MAE & MSE  & MAE & MSE & MAE& MSE & MAE& MSE  & MAE& MSE  & MAE& MSE  & MAE& MSE  & MAE\\
\midrule

\multirow{5}{*}{\rotatebox{90}{$Weather$}}
& 96  &0.155&0.201& 0.163 & 0.215 & 0.190 & 0.240 & 0.171 & 0.224 & 0.165 & 0.215 & 0.184& 0.230& 0.188 & 0.253 & 0.221 & 0.297  \\
& 192 &0.202&0.245& 0.210 & 0.254 & 0.243 & 0.284 & 0.215 & 0.263 & 0.210 & 0.257 & 0.245& 0.283& 0.250 & 0.304 & 0.270 & 0.322 \\
& 336 &0.251&0.281& 0.256 & 0.292 & 0.270 & 0.305 & 0.258 & 0.299 & 0.259 & 0.297 & 0.305& 0.321& 0.312 & 0.346 & 0.320 & 0.351  \\
& 720 &0.320&0.329& 0.321 & 0.339 & 0.348 & 0.359 & 0.320 & 0.346 & 0.332 & 0.346 & 0.381& 0.371& 0.387 & 0.393 & 0.390 & 0.396 \\
&Avg.&{\bf0.232}&{\bf0.264}&\underline{0.238} &\underline{0.275} &0.263 &0.297 &0.241 &0.283 &0.242 &0.279 &0.279 &0.301 &0.284 &0.324 &0.300 &0.342 \\
\midrule

\multirow{5}{*}{\rotatebox{90}{$ETTh1$}}
& 96  &0.439&0.447& 0.458 & 0.456 & 0.601 & 0.536 & 0.492 & 0.495 & 0.516 & 0.485 & 0.861& 0.628& 0.512 & 0.499 & 0.613 & 0.552 \\
& 192 &0.556&0.515& 0.570 & 0.516 & 0.709 & 0.587 & 0.565 & 0.538 & 0.598 & 0.524 & 0.797& 0.593& 0.624 & 0.555 & 0.722 & 0.598 \\
& 336 &0.601&0.543& 0.608 & 0.535 & 0.801 & 0.635 & 0.721 & 0.622 & 0.657 & 0.550 & 0.941& 0.648& 0.691 & 0.574 & 0.750 & 0.619 \\
& 720 &0.708&0.581& 0.725 & 0.591 & 1.385 & 0.831 & 0.986 & 0.743 & 0.762 & 0.610 &0.877 &0.641 & 0.728 & 0.614 & 0.721 & 0.616 \\
&Avg.&{\bf0.576}&{\bf0.429}&\underline{0.590} &\underline{0.525} &0.874 &0.647 &0.691 &0.600 &0.633 &0.542 &0.869 &0.628 &0.639 &0.561 &0.702 &0.596 \\
\midrule

\multirow{5}{*}{\rotatebox{90}{$ETTh2$}}
& 96  &0.306&0.357& 0.331 & 0.374 & 0.539 & 0.495 & 0.357 & 0.411 & 0.353 & 0.389 & 0.378& 0.409& 0.382 & 0.416 & 0.413 & 0.451 \\
& 192 &0.378&0.399& 0.402 & 0.411 & 0.675 & 0.555 & 0.569 & 0.519 & 0.403 & 0.414 & 0.490& 0.467& 0.478 & 0.474 & 0.474 & 0.477 \\
& 336 &0.382&0.417& 0.406 & 0.433 & 0.718 & 0.580 & 0.671 & 0.572 & 0.426 & 0.441 & 0.537& 0.494& 0.504 & 0.501 & 0.547 & 0.543 \\
& 720 &0.435&0.456& 0.449 & 0.464 & 0.732 & 0.605 & 0.824 & 0.648 & 0.477 & 0.480 & 0.510& 0.491& 0.499 & 0.509 & 0.516 & 0.523 \\
&Avg.&{\bf0.375}&{\bf0.407}&\underline{0.397} &\underline{0.421} &0.666 &0.559 &0.605 &0.538 &0.415 &0.431 &0.479 &0.465 &0.466 &0.475 &0.488 &0.499 \\
\midrule

\multirow{5}{*}{\rotatebox{90}{$ETTm1$}}
& 96  &0.370&0.397& 0.390 & 0.404 & 0.610 & 0.508 & 0.352 & 0.392 & 0.410 & 0.419 & 0.583& 0.501& 0.578 & 0.518 & 0.774 & 0.614 \\
& 192 &0.419&0.429& 0.429 & 0.423 & 0.666 & 0.540 & 0.382 & 0.412 & 0.437 & 0.434 & 0.630& 0.528& 0.617 & 0.546 & 0.754 & 0.592 \\
& 336 &0.466&0.452& 0.469 & 0.439 & 0.895 & 0.615 & 0.419 & 0.434 & 0.476 & 0.454 & 0.725& 0.568& 0.998 & 0.775 & 0.869 & 0.677  \\
& 720 &0.564&0.510& 0.569 & 0.498 & 0.916 & 0.646 & 0.490 & 0.477 & 0.681 & 0.556 & 0.769& 0.549& 0.693 & 0.579 & 0.810 & 0.630  \\
&Avg.&\underline{0.454}&0.447&0.464 &\underline{0.441} &0.772 &0.577 &\textbf{0.411} &\textbf{0.429} &0.501 &0.466 &0.677 &0.537 &0.722 &0.605 &0.802 &0.628 \\
\midrule

\multirow{5}{*}{\rotatebox{90}{$ETTm2$}}
& 96  &0.186&0.265& 0.188 & 0.269 & 0.283 & 0.344 & 0.213 & 0.303 & 0.191 & 0.274 & 0.212& 0.285& 0.291 & 0.399 & 0.352 & 0.454\\
& 192 &0.250&0.309& 0.251 & 0.309 & 0.353 & 0.384 & 0.278 & 0.345 & 0.252 & 0.317 & 0.270& 0.323& 0.307 & 0.379 & 0.694 & 0.691 \\
& 336 &0.302&0.342& 0.307 & 0.346 & 0.420 & 0.422 & 0.338 & 0.385 & 0.306 & 0.353 & 0.323& 0.353& 0.543 & 0.559 & 2.408 & 1.407 \\
& 720 &0.413&0.408 & 0.426 & 0.417 & 0.553 & 0.491 & 0.436 & 0.440 & 0.433 & 0.427 & 0.474& 0.449& 0.712 & 0.614 & 1.913 & 1.166 \\
&Avg.&{\bf0.287}&{\bf0.331}&\underline{0.293} &\underline{0.335} &0.402 &0.410 &0.316 &0.368 &0.296 &0.343 &0.320 &0.353 &0.463 &0.488 &1.342 &0.930\\
\midrule


\multirow{5}{*}{\rotatebox{90}{$ECL$}}
& 96  &0.139&0.234& 0.139 & 0.237 & 0.142 & 0.240 & 0.150 & 0.253 & 0.140 & 0.238 & 0.299& 0.373& 0.231 & 0.323 & 0.261 & 0.348\\
& 192 &0.158&0.251& 0.156 & 0.252 & 0.158 & 0.254 & 0.164 & 0.264 & 0.160 & 0.255 & 0.305& 0.379& 0.261 & 0.356 & 0.338 & 0.406\\
& 336 &0.182&0.277& 0.175 & 0.270 & 0.175 & 0.271 & 0.181 & 0.282 & 0.180 & 0.276 &0.319 &0.391 & 0.360 & 0.445 & 0.410 & 0.474 \\
& 720 &0.247&0.331& 0.233 & 0.317 & 0.230 & 0.315 & 0.223 & 0.321 & 0.241 & 0.323 & 0.369& 0.426& 0.530 & 0.585 & 0.715 & 0.685 \\
&Avg.&0.181&0.273&\textbf{0.176} &\textbf{0.269} &\textbf{0.176} &\underline{0.270} &0.180 &0.280 &0.180 &0.273 &0.323 &0.392 &0.346 &0.427 &0.431 &0.478\\
\midrule

\multirow{5}{*}{\rotatebox{90}{$Traffic$}}
& 96  &0.416&0.283& 0.414 & 0.297 & 0.478 & 0.368 & 0.419 & 0.298 & 0.403 & 0.289 & 0.719& 0.416& 0.639 & 0.400 & 0.672 & 0.405\\
& 192 &0.424&0.292& 0.426 & 0.301 & 0.481 & 0.363 & 0.434 & 0.305 & 0.415 & 0.296 & 0.748& 0.428& 0.637 & 0.416 & 0.727 & 0.424\\
& 336 &0.432&0.297& 0.434 & 0.303 & 0.488 & 0.365 & 0.449 & 0.313 & 0.426 & 0.304 & 0.853& 0.471& 0.655 & 0.427 & 0.749 & 0.454 \\
& 720 &0.480&0.335& 0.487 & 0.337 & 0.537 & 0.386 & 0.484 & 0.336 & 0.474 & 0.331 & 1.485& 0.825& 0.722 & 0.456 & 0.847 & 0.499\\
&Avg.&\underline{0.438}&{\bf0.301}&0.440 &0.310 &0.496 &0.371 &0.447 &0.313 &\textbf{0.430} &\underline{0.305} &0.951 &0.535 &0.663 &0.425 &0.749 &0.446\\
\midrule
\multicolumn{2}{c|}{Average}&
{\bf0.363}&{\bf0.350}& \underline{0.371} &\underline{0.367} &0.521 &0.447 &0.413 &0.401 &0.385 &0.376 &0.556 &0.458 &0.511 &0.472 &0.687 &0.559  \\

\bottomrule
\end{tabular}%
}
\end{center}
\end{table*}
\newpage
\clearpage  

\end{document}